\PassOptionsToPackage{twoside=false}{geometry}
\PassOptionsToPackage{table,RGB}{xcolor}
\documentclass[manuscript]{acmart}

\setcopyright{acmlicensed}
\acmJournal{CSUR}
\AtBeginDocument{%
  }

\renewcommand\footnotetextcopyrightpermission[1]{}

\usepackage{mathtools}
\usepackage{amsthm}
\usepackage{bm}
\usepackage{graphicx}
\usepackage{booktabs}
\usepackage{multirow}
\usepackage{array}
\newcolumntype{C}[1]{>{\centering\arraybackslash}m{#1}}
\newcolumntype{L}[1]{>{\raggedright\arraybackslash}m{#1}}
\usepackage{tabularx}
\usepackage{tabulary}
\usepackage{makecell}
\usepackage{diagbox}
\usepackage{adjustbox}
\usepackage{float}
\usepackage{enumitem}
\usepackage{xspace}
\usepackage{xurl}
\usepackage{url}
\usepackage[capitalize,noabbrev]{cleveref}
\usepackage{algorithm}
\usepackage{algorithmic}
\usepackage{dsfont}
\usepackage{pifont}
\usepackage{bbding}
\usepackage{fancybox}
\usepackage{tikz}
\usetikzlibrary{positioning,calc,fit,backgrounds,arrows.meta}
\usepackage[edges]{forest}
\usepackage{xcolor}
\usepackage{colortbl}

\definecolor{Gray}{gray}{0.9}
\definecolor{hidden-draw}{RGB}{20,68,106}
\definecolor{hidden-pink}{RGB}{255,245,247}
\definecolor{lightgrey}{gray}{0.92}
\definecolor{light_double_grey}{gray}{0.95}
\definecolor{lightred}{RGB}{251,49,153}
\definecolor{lightorange}{RGB}{244,230,217}
\definecolor{LightRed}{rgb}{1,0.92,0.92}
\definecolor{LightBlue}{rgb}{0.9,0.94,1}
\definecolor{LightGreen}{rgb}{0.9,1.0,0.88}

\definecolor{BoxFill}{HTML}{F8F4E8}
\definecolor{SubColor}{HTML}{3E2C1C}
\definecolor{rootcolor}{RGB}{250,200,200}
\definecolor{dimcolor}{RGB}{215,200,240}
\definecolor{dataipcolor}{RGB}{200,225,250}
\definecolor{modelipcolor}{RGB}{200,240,210}
\definecolor{catdatacolor}{RGB}{218,236,255}
\definecolor{catmodelcolor}{RGB}{218,246,226}
\definecolor{memdatacolor}{RGB}{235,245,255}
\definecolor{memmodelcolor}{RGB}{235,250,240}

\tikzset{
  card/.style={
    draw=black!12,
    rounded corners=10pt,
    inner xsep=12pt,
    inner ysep=9pt,
    align=left,
    text=black,
    fill=BoxFill
  },
  groupbox/.style={
    draw=black!15,
    rounded corners=14pt
  },
  L1/.style=card, L2/.style=card, L3/.style=card,
  R1/.style=card, R2/.style=card, R3/.style=card, R4/.style=card, R5/.style=card,
  treeedge/.style={
    -{Stealth[length=1.5mm, width=1mm]},
    thin,
    draw
  }
}

\newcommand{\seclink}[2]{\hyperref[#1]{\textbf{#2}}}
\newcommand{\subtxt}[1]{\textcolor{SubColor}{\footnotesize #1}}
\newcommand{\sublink}[2]{\subtxt{\hyperref[#1]{\textbf{#2}}}}

\title[IP Protection in the Era of Visual Generative AI: A Survey]{IP Protection in the Era of Visual Generative AI: A Survey}

\author{Zhuan Shi}
\email{zhuan.shi@mila.quebec}
\affiliation{
  \institution{Mila - Quebec AI Institute and McGill University}
  \city{Montreal}
  \country{Canada}
}

\author{Shunchang Liu}
\email{shunchang.liu@epfl.ch}
\affiliation{
  \institution{EPFL}
  \city{Lausanne}
  \country{Switzerland}
}

\author{Alireza Dehghanpour Farashah}
\email{alireza.farashah@mila.quebec}
\affiliation{
  \institution{Mila - Quebec AI Institute and McGill University}
  \city{Montreal}
  \country{Canada}
}

\author{Qian Yang}
\email{qian.yang@mila.quebec}
\affiliation{
  \institution{Mila - Quebec AI Institute and Université de Montréal}
  \city{Montreal}
  \country{Canada}
}

\author{Han Yu}
\email{han.yu@ntu.edu.sg}
\affiliation{
  \institution{Nanyang Technological University}
  \city{Singapore}
  \country{Singapore}
}

\author{Cao Yang}
\email{cao@c.titech.ac.jp}
\affiliation{
  \institution{Science Tokyo}
  \city{Tokyo}
  \country{Japan}
}

\author{Chaochao Chen}
\email{zjuccc@zju.edu.cn}
\affiliation{
  \institution{Zhejiang University}
  \city{Hang Zhou}
  \country{China}
}
\author{Yuping Yan}
\email{yanyuping@westlake.edu.cn}
\affiliation{
  \institution{Westlake University}
  \city{Hang Zhou}
  \country{China}
}

\author{Yaochu Jin}
\email{jinyaochu@westlake.edu.cn}
\affiliation{
  \institution{Westlake University}
  \city{Hang Zhou}
  \country{China}
}

\author{Golnoosh Farnadi}
\email{farnadig@mila.quebec}
\affiliation{
  \institution{Mila - Quebec AI Institute and McGill University}
  \city{Montreal}
  \country{Canada}
}

\author{Lingjuan Lyu}
\email{lingjuan.lv@sony.com}
\affiliation{
  \institution{Sony Research}
  \city{Zurich}
  \country{Switzerland}
}

\renewcommand{\shortauthors}{Shi et al.}

\begin{document}

\begin{abstract}

The rapid evolution of visual generative AI has introduced a wide range of intellectual property risks, spanning the unauthorized learning, reproduction, extraction, misuse, and redistribution of protected data and model assets. To address these risks, a growing body of technical defenses has been proposed. However, existing surveys typically organize this literature by lifecycle stage or technical mechanism, which can obscure the protective intent of different methods. This survey presents a two-dimensional taxonomy for IP protection in visual generative models. The primary axis is a \textit{Control Logic View}, which classifies methods into \textit{Information Exposure Control}, \textit{Generative Behavior Constraint}, and \textit{Attribution \& Accountability} according to the risk variable they regulate. The secondary axis distinguishes \textit{Data IP} from \textit{Model IP} as cross-cutting asset dimensions. Under this framework, we systematically review protection methods, align evaluation protocols with protection objectives, and discuss open challenges including proactive model-level safeguards, standardized evaluation, robustness against adaptive attacks, and explainable evidence. This survey aims to offer a principled, systematic, and easy-to-follow overview for both new and experienced researchers in visual generative AI IP protection.
\end{abstract}


\keywords{visual generative AI, intellectual property protection, copyright protection}

\maketitle

\section{Introduction}
\label{sec:intro}

\begin{figure}[t]
    \centering
    \includegraphics[width=0.9\linewidth]{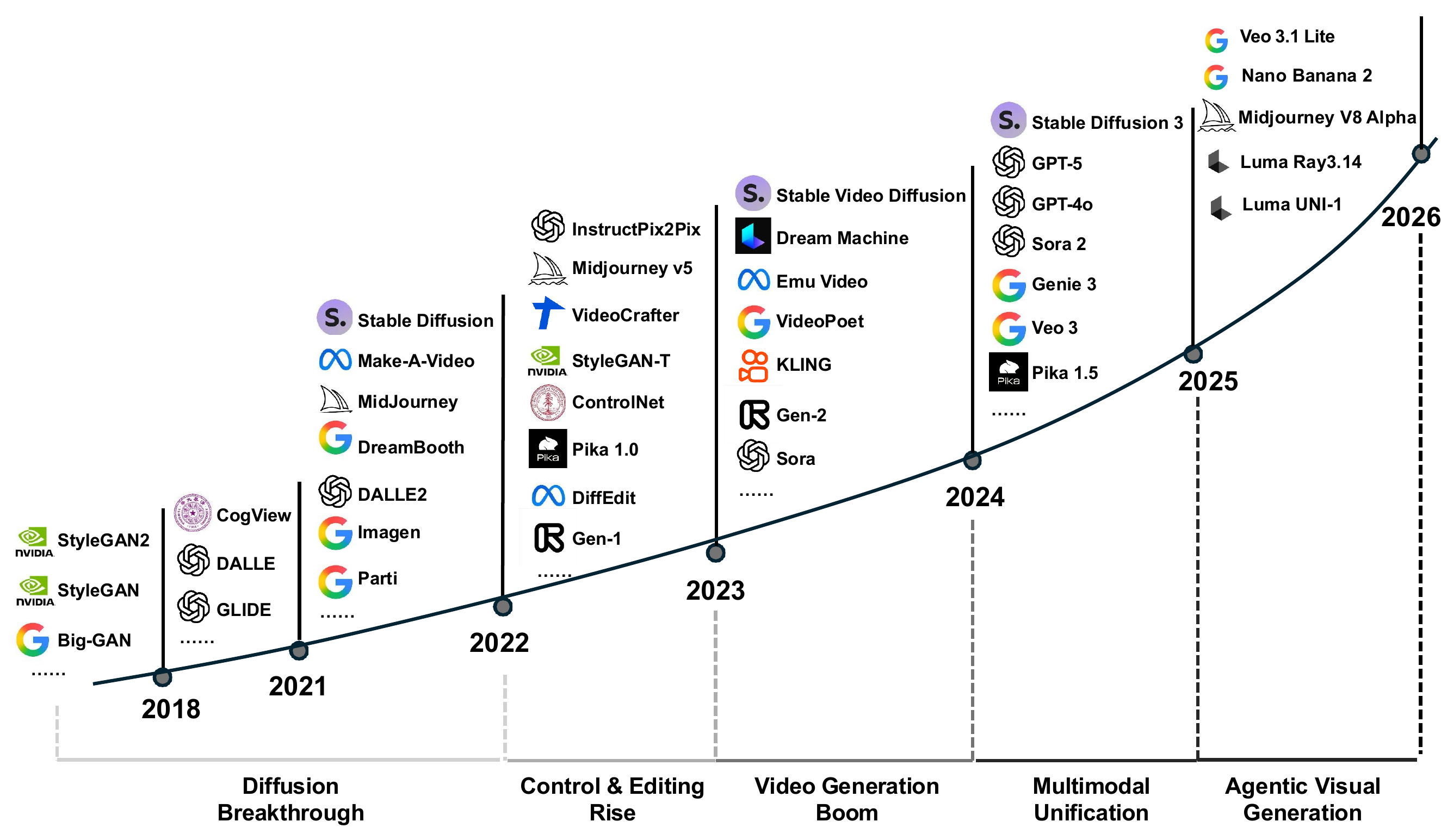}
    \caption{The emergence of notable visual generative models over time (2018-2026).}
    \Description{A timeline from 2018 to 2026 showing representative visual generative models and major development stages, including diffusion breakthrough, control and editing, video generation, multimodal unification, and agentic visual generation.}
    \label{fig:dm}
\end{figure}

The rapid advancement of visual generative artificial intelligence (AI) models has fundamentally reshaped the landscape of AI-generated content (AIGC), enabling the creation of images and videos with unprecedented realism, diversity, and controllability~\cite{openai2024,betker2023improving,rombach2022high,brooks2024video}. 
This technological breakthrough has given rise to a new class of widely adopted systems (Figure~\ref{fig:dm}), including OpenAI's DALL·E 3~\cite{betker2023improving} and Sora 2~\cite{openai2025sora2}, Pika~\cite{pika}, Midjourney~\cite{midjourney2023}, Stability AI's Stable Diffusion 3~\cite{esser2024scaling} and Stable Video Diffusion~\cite{svd}, as well as Google DeepMind's Imagen~\cite{imagen}, Veo 3~\cite{google2025veo3}, and the interactive world model Genie 3~\cite{deepmind2025genie3}. These models have been rapidly integrated into commercial products and creative workflows, accelerating transformative changes across digital art, media, design, advertising, entertainment, education, and healthcare. As time, cost and specialized skill barriers continue to decrease, visual generative AI is starting to show significant economic and cultural impact.

The progress of generative AI is fundamentally driven by access to large-scale, high-quality data, which enables researchers and developers to train, evaluate, and refine increasingly capable models~\cite{li2025rethinkingdataprotectiongenerative}. For example, modern multimodal foundation models such as GPT-5~\cite{openai2025gpt5} and GPT-4o~\cite{openai2024gpt4ocard} are trained on diverse sources that might include publicly available datasets, data crawled from the web, licensed archives, and metadata. Similarly, advanced video generation systems, such as Veo 3~\cite{google2025veo3}, rely on vast collections of visual and audiovisual data to learn realistic motion, physical consistency, and synchronized audio generation. This data dependency is the central reason for intellectual property (IP) protection to become increasingly urgent: \textit{the success of visual generative AI relies on the acquisition, processing, and learning from valuable creative assets}.

Beyond the training data, both trained models and their outputs are themselves valuable assets. For model developers and vendors, model weights, architectures, training recipes, aligned behaviors, and transferable add-ons (e.g., fine-tuned checkpoints) represent substantial investment and are often commercialized through APIs, licenses, and closed deployment services. For downstream users, generated images and videos are increasingly used as production-ready contents in creative and industrial pipelines, meaning that value is also realized at inference and deployment stages. As visual generation becomes both economically consequential and massively scalable, IP risks extend well beyond data provenance. It now encompasses the protection of training data and the models as well as the accountability of generated outputs.

These developments underscore why IP protection in visual generative AI is both urgent and fundamentally different from traditional software IP protection. Infringement might arise from the unauthorized use of training data, the extraction or theft of proprietary model information, or the generation of outputs that reproduce proprietary contents, styles, or identities. Accordingly, the risk landscape cannot be understood through a single lens. In this survey, we frame the problem through two complementary asset dimensions: \textit{Data IP} and \textit{Model IP}.

\begin{itemize}
    \item \textbf{Data IP concerns} arise from the unauthorized exploitation of protected training materials and creative content. As visual generative models require massive datasets, they might take in copyrighted artworks, branded assets, character images, or identity-bearing content without consent or appropriate licensing. As a result, infringement is not limited to direct copying or memorization, it also extends to near-duplicate reproduction, style mimicry, and high-similarity generation that could dilute the market value and creative rights of the original contents.
    
    \item \textbf{Model IP concerns} relate to the protection of the generative AI systems themselves. State-of-the-art visual generative models require substantial computational resources, engineering expertise, data curation, and alignment effort, making their weights, architectures, behaviors, and capabilities valuable IP assets. These assets are vulnerable not only to direct theft or unauthorized redistribution, but also to distillation, model extraction, and capability replication, where an adversary attempts to reproduce the functionality or commercial value of a proprietary model without bearing proper development costs.
\end{itemize}

Although a growing body of technical work has begun to address these risks, the landscape of IP protection in visual generative AI remains highly fragmented. Existing surveys and reviews typically organize methods either by \textit{lifecycle stage} (e.g., training, fine-tuning, or inference) or by \textit{technical mechanism} (e.g., watermarking, machine unlearning, or adversarial perturbation). However, both views have important limitations which makes it challenging for newcomers to the field to gain in-depth understanding quickly. A lifecycle-based taxonomy groups together methods that are deployed at the same stage even when they intervene on fundamentally different IP risk variables. For instance, training-time methods may either suppress the model’s retention of protected content, or deliberately inject provenance signals such as watermarks to facilitate later ownership verification and misuse tracing. These methods are not unified by protective intent. They merely happen to be applied at the same phase. A mechanism-based taxonomy, by contrast, emphasizes \emph{how} a method operates rather than \emph{what} risk it is designed to regulate. Consequently, existing categorizations often blur the underlying control objective and leave a fundamental question insufficiently addressed: \emph{which specific IP risk variable is the method actually controlling?}

To address this gap, we organize this survey under a \textit{Control Logic View}. Rather than classifying methods by what stage they are applied, we categorize them by their primary \textit{control objective}, namely, the specific IP risk they are designed to regulate. Under this view, the diverse landscape of visual IP protection methods can be unified into three fundamental control objectives:
\begin{itemize}
    \item \textbf{Information Exposure Control.} This objective controls what protected assets can be internalized or revealed by the model. From the Data IP perspective, the goal is to reduce memorization and unauthorized learning from protected samples by intervening at the level of data exposure or learnability, for example through filtering, corruption, compositional training, or perturbation. From the Model IP perspective, the goal is to restrict unauthorized access to proprietary model value, for example through model locking or authorization-dependent usability control.
    
    \item \textbf{Generative Behavior Constraint.} This objective controls how protected content or model capabilities can be reproduced, expressed, or misused during generation. For Data IP, it includes prompt mediation, concept unlearning, and inference-time guidance that reduce the generation of infringing styles, objects, or identities. For Model IP, it includes defenses against model extraction and unauthorized personalization, which aim to prevent capability transfer or misuse through querying, distillation, editing, or fine-tuning.
    
    \item \textbf{Attribution \& Accountability.} This objective controls whether ownership or misuse can be verified, attributed, and proven after deployment. For Data IP, this includes data usage verification and data watermarking for traceability. For Model IP, this includes origin attribution, fingerprinting, and watermarking mechanisms that support ownership verification, model lineage analysis, and evidence-based tracing.
\end{itemize}

Under each control objective, we further distinguish between \textit{Data IP} and \textit{Model IP} protection. In this taxonomy, Table~I serves as the primary organization of protection methods, while Table~II mirrors the same structure for evaluation protocols. Our survey contributes not only a unified taxonomy of protection methods, but also a consistent evaluation view aligned with the same control logic. More broadly, these three objectives reflect different intervention logic, ranging from upstream prevention of exposure, to downstream constraint of generation behavior, to post-hoc verification and accountability.
By organizing the literature in this way, we aim to provide a more principled and function-oriented understanding of visual IP protection. This perspective makes it easier to compare methods that may look different technically but regulate similar risk variables, and it also helps reveal structural gaps in the current landscape, such as the shortage of proactive model-level defenses, the lack of standardized evaluation, and the need for more explainable and architecture-agnostic protection mechanisms. The main contributions of this survey are as follows:
\begin{itemize}
    \item We propose a two-dimensional taxonomy for IP protection in visual generative AI, where an intent-driven \textit{Control Logic View} serves as the primary axis and \textit{Data IP} versus \textit{Model IP} serves as a cross-cutting asset dimension.
    \item We systematically review protection methods and evaluation protocols under the same control-logic framework, enabling consistent comparison of effectiveness, robustness, and resilience against adaptive attacks across heterogeneous techniques.
    \item We bridge technical methods with regulatory and industrial practices, and identify open challenges such as proactive model-level safeguards, standardized benchmarking, explainable attribution, robustness to adaptive misuse, and architecture-agnostic protection.

\end{itemize}

The rest of this paper is organized as follows. Section~\ref{sec:background} introduces the necessary background on visual generative models and frames IP infringement through the complementary lenses of Data IP and Model IP. Section~\ref{sec:IPP} presents our Control Logic View and systematically reviews technical protection methods under three control objectives: (i) Information Exposure Control, (ii) Generative Behavior Constraint, and (iii) Attribution \& Accountability, each further divided into Data IP protection and Model IP protection (Table.\ref{table:control_logic_view}). Section~\ref{sec:eval} summarizes evaluation protocols and metrics organized under the same control-logic framework, with an emphasis on both effectiveness and robustness (Table.\ref{table:metric}). Section~\ref{sec:reg} surveys legal and regulatory developments, as well as industry participation relevant to IP protection in visual generative AI. Section~\ref{sec:future} discusses open challenges and future directions toward standardized, proactive, explainable, and architecture-agnostic IP defense mechanisms. Finally, Section~\ref{sec:concl} concludes the survey.

\begin{figure*}[t]
    \centering
    \includegraphics[width=1.0\textwidth]{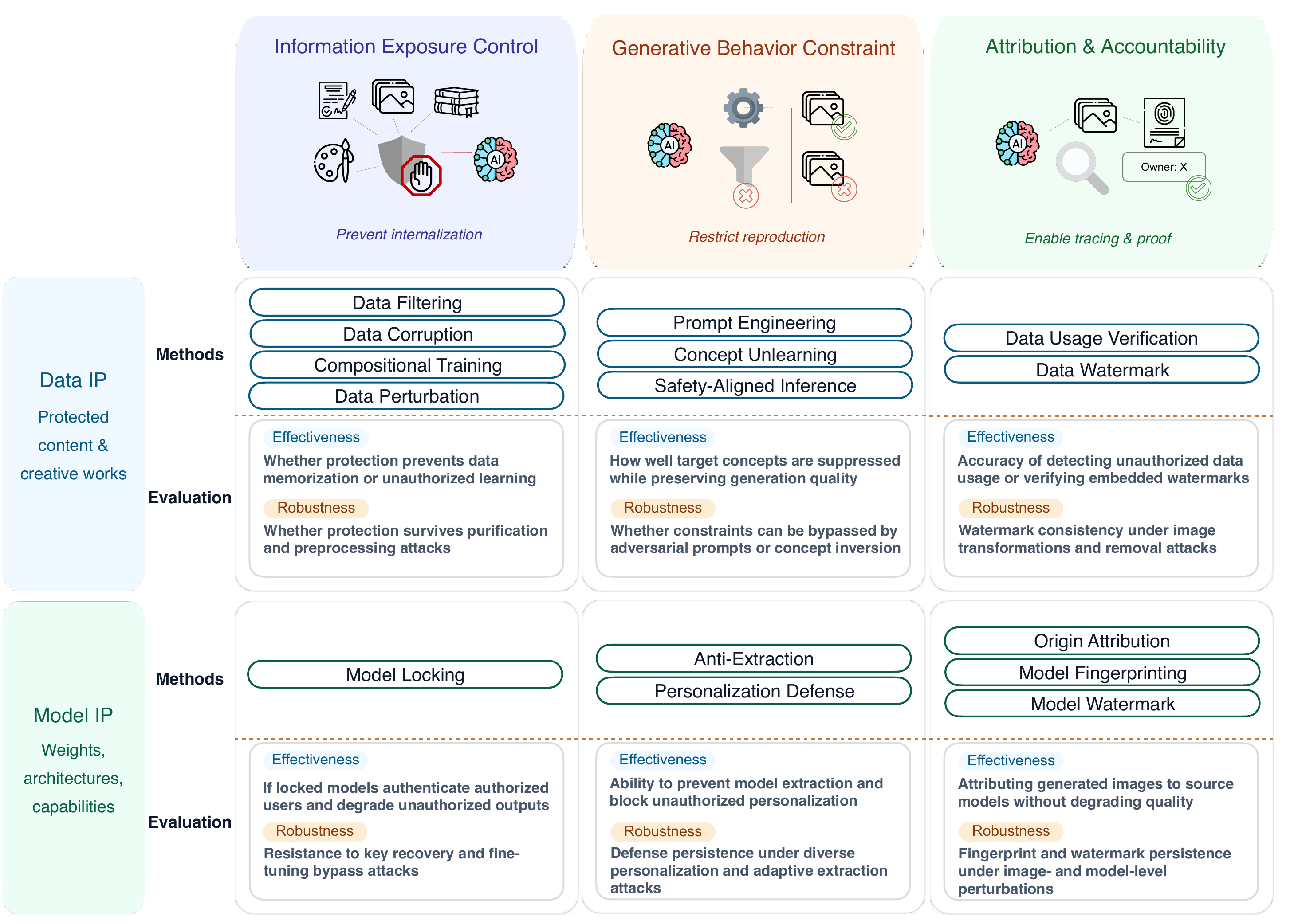}
    \caption{\textbf{Taxonomy of IP protection methods for visual generative models based on the Control Logic View.}}
    \Description{A two-dimensional taxonomy of IP protection methods for visual generative models. The columns represent three control objectives: Information Exposure Control, Generative Behavior Constraint, and Attribution and Accountability. The rows distinguish Data IP and Model IP. Each cell lists representative protection method families and evaluation dimensions.}
    \label{fig:taxonomy}
\end{figure*}

\section{Background}
\label{sec:background}

This section provides the background necessary for understanding the IP protection landscape of visual generative AI. Rather than offering an exhaustive tutorial on generative modeling, we focus on the architectural and problem-level context that is most relevant to the taxonomy developed later in this survey. Specifically, we first review major families of visual generative models with an emphasis on the design properties that shape IP risks and protection opportunities. We then characterize IP infringement through the complementary lenses of \textit{Data IP} and \textit{Model IP}, which serve as the asset-level foundation for the control-oriented taxonomy introduced in Section~\ref{sec:IPP}.

\subsection{Visual Generative Models}
\label{sec:back01}

Visual generative models learn to synthesize images or videos by approximating complex data distributions from large-scale training corpora. Although these models differ substantially in architecture and sampling process, their design choices directly influence what information may be internalized, how generation can be controlled, and where IP protection mechanisms can intervene. Different architectures expose protection-relevant information through different internal interfaces. Variational Autoencoders (VAEs) expose compressed latent codes; Generative Adversarial Networks (GANs) expose learned latent manifolds and generator--discriminator interactions; autoregressive models expose explicit token-by-token conditional distributions; and diffusion-based models expose iterative denoising trajectories together with rich conditioning hooks such as text embeddings, cross-attention, and classifier-free guidance. These architectural interfaces are highly relevant to the protection methods reviewed later, particularly for memorization mitigation, concept removal, behavior constraint, watermarking, and attribution.

\textbf{Variational Autoencoders (VAEs).}
VAEs~\cite{kingma2013auto} are probabilistic latent-variable models that learn to encode data into a compressed latent space and decode latent variables back into observations. Their standard objective maximizes the evidence lower bound:
\begin{equation}
\label{eq:vae}
\begin{aligned}
L(\phi,\theta)
= \mathbb{E}_{z \sim q(z \mid x;\phi)}[\log p(x \mid z;\theta)]
- \mathrm{KL}(q(z \mid x;\phi)\|p(z)).
\end{aligned}
\end{equation}
The reconstruction term encourages faithful generation, while the KL term regularizes the latent distribution toward a prior. VAEs have been widely used for image generation, representation learning, data compression, and anomaly detection~\cite{kulkarni2015deep,liu2018image,deshpande2017learning,higgins2016beta,hu2023glso}. Numerous variants, including conditional VAEs, mixture-based VAEs, and adversarially enhanced VAEs, have been proposed to improve fidelity and controllability~\cite{sohn2015learning,dilokthanakul2016deep,mescheder2017adversarial,kim2021conditional,silvestri2022deterministic,bae2022multi}.

From the perspective of IP protection, VAEs are important because they expose an early and influential form of \emph{representation-space compression}. In such models, protected information is not necessarily memorized, reproduced, or leaked directly in pixel space; instead, it may be encoded into structured latent variables that preserve semantic, stylistic, or instance-specific information. This changes the nature of IP risk: infringement is no longer limited to pixel-level copying or near-duplicate reproduction, but may also emerge through representation-level vulnerabilities, including latent inversion, membership inference over latent codes, and latent-space manipulation. Through these mechanisms, an adversary may recover protected attributes, infer whether protected samples influenced training, or recombine latent factors in ways that reproduce protected visual characteristics. Although VAEs are no longer the dominant paradigm for state-of-the-art visual generation, the notion of latent bottlenecks remains central to modern generative systems, especially latent diffusion models. Therefore, VAEs provide useful background for understanding how protected information can be internalized, transformed, and partially recoverable from learned latent representations.

\textbf{Generative Adversarial Networks (GANs).}
GANs~\cite{goodfellow2020generative} generate data through adversarial training between a generator $G$ and a discriminator $D$. A standard GAN objective is:
\begin{equation}
\label{eq:gan}
\begin{aligned}
L(\theta,\phi)
&=
\min_{\theta}\max_{\phi}
\mathbb{E}_{x \sim p_{\mathrm{data}}}[\log D(x;\phi)] \\
&\quad + \mathbb{E}_{\hat{x}\sim G(\theta)}[\log(1-D(\hat{x};\phi))].
\end{aligned}
\end{equation}
GANs have been widely used for image synthesis, style transfer, face generation, and image-to-image translation~\cite{goodfellow2020generative,zhu2017unpaired,ho2016generative}. Over time, variants such as WGAN, CycleGAN, Progressive GAN, and StyleGAN improved training stability and visual fidelity~\cite{arjovsky2017wasserstein,gulrajani2017improved,zhu2017unpaired,karras2017progressive,karras2019style,karras2020analyzing,karras2021alias}.

From the perspective of IP protection, GANs are important because they expose a different set of protection-relevant interfaces from VAEs. Instead of relying primarily on compressed latent codes, GANs organize generation through a learned latent manifold and an adversarial generator--discriminator interaction. This makes them relevant to IP risks such as style imitation, identity synthesis, memorization, and model attribution. The latent manifold can encode semantically meaningful directions that may be manipulated to reproduce protected visual attributes, while the discriminator and generated-output distributions can reveal model-specific artifacts useful for fingerprinting and source attribution. Many early studies on model stealing, generative fingerprints, and ownership verification were first developed in GAN settings, and these ideas later influenced attribution and watermarking methods for diffusion-based visual generators.

\textbf{Autoregressive Models (ARMs).}
Autoregressive models generate visual content sequentially by factorizing the joint distribution over pixels, patches, or discrete tokens:
\begin{equation}
\label{eq:arm}
p(\mathbf{x}) = \prod_{i=1}^{n} p(x_i \mid x_1,\ldots,x_{i-1};\theta).
\end{equation}
Early approaches such as PixelCNN and PixelRNN generated images pixel-by-pixel~\cite{van2016conditional,van2016pixel}. Later work improved scalability by decoupling representation learning from sequence modeling: VQ-VAE and VQ-VAE-2 first encode images into compact vector-quantized latent tokens, enabling autoregressive models to model discrete latent sequences rather than raw pixels~\cite{van2017neural,razavi2019generating}. This design was further extended to large-scale text-to-image generation by transformer-based models such as DALL-E~\cite{ramesh2021zero}. More recent advances, including VAR and Infinity, have substantially improved the efficiency and quality of visual autoregressive generation~\cite{tian2024visual,han2025infinity}. Multimodal systems such as GPT-4o and GPT-5 also illustrate the growing importance of autoregressive reasoning in integrated text-image generation pipelines~\cite{openai2024gpt4ocard,openai2025gpt5}.

From an IP perspective, autoregressive models are particularly important because they expose generation as an explicit sequence of conditional decisions. This makes them conceptually relevant for studying memorization, verbatim reproduction, and behavior imitation at the token or sequence level. Compared with diffusion models, visual ARMs have received far less attention in the IP protection literature, yet they may present distinct vulnerabilities and opportunities, especially for attribution, extraction analysis, and output-level similarity control. Their relative underexploration also motivates one of the future directions highlighted in Section~\ref{sec:future}.

\textbf{Diffusion Models and Flow-Based Methods.}
Diffusion models, introduced by Sohl-Dickstein et al.~\cite{sohl2015deep}, generate data by learning to reverse a progressive noising process. In the forward process, Gaussian noise is gradually added to data:
\begin{equation}
\label{eq:diffusion_forward}
q(\mathbf{x}_t \mid \mathbf{x}_{t-1})
=
\mathcal{N}\!\left(
\mathbf{x}_t;
\sqrt{1-\beta_t}\mathbf{x}_{t-1},
\beta_t \mathbf{I}
\right),
\end{equation}
where $\beta_t$ controls the variance schedule. Sampling then proceeds by learning a reverse process that denoises step by step:
\begin{equation}
\label{eq:diffusion_reverse}
p_{\theta}(\mathbf{x}_{t-1}\mid \mathbf{x}_t), \quad t=T,T-1,\ldots,1.
\end{equation}
Diffusion models have demonstrated outstanding fidelity and diversity in image synthesis and have become the dominant paradigm for modern text-to-image and text-to-video generation~\cite{song2019generative,dhariwal2021diffusion,ho2020denoising,song2020score}.

Closely related are flow-based generative models, including normalizing flows and more recent flow-matching formulations~\cite{rezende2015variational,lipman2022flow,liu2022flow,albergo2022building}. A normalizing flow maps a simple prior variable $z_0$ to data $x$ through a sequence of invertible transformations:
\begin{equation}
\label{eq:flow}
x = f_K\!\left(f_{K-1}\!\left(\cdots f_0(z_0)\right)\right),
\qquad z_0 \sim \mathcal{N}(0,I).
\end{equation}
Flow matching provides a simulation-free training paradigm for continuous normalizing flows, and diffusion models can be interpreted as a special case under Gaussian probability paths~\cite{lipman2022flow}. Rectified flow further improves efficiency by encouraging straighter generation trajectories, and this connection underlies recent models such as Stable Diffusion 3~\cite{liu2022flow,esser2024scaling}.

Diffusion-based visual generation has advanced rapidly in both image and video synthesis. Stable Diffusion~\cite{rombach2022high} popularized latent diffusion, which greatly reduced computational cost by operating in a compressed latent space. Stable Diffusion 3~\cite{esser2024scaling} introduced the MM-DiT architecture with rectified flow. DALL·E 3~\cite{betker2023improving}, Imagen 2~\cite{deepmind2023imagen2}, and Midjourney~\cite{midjourney2023} further improved text understanding and generation quality. In video, Sora, Sora 2, Veo 3, and CogVideoX have substantially expanded the realism, duration, and multimodal richness of generated videos~\cite{openai2024sora,openai2025sora2,google2025veo3,yang2024cogvideox}. Interactive world models such as Genie 3 point to a further shift from static media generation to persistent, navigable, and behaviorally rich environments~\cite{deepmind2025genie3}.

Diffusion and flow-based models are central to this survey because most current IP protection methods for visual generative AI are developed in this setting. Their importance comes not only from their empirical dominance, but also from the fact that they expose multiple control interfaces that are highly amenable to protection: training data composition, denoising trajectories, latent representations, conditioning embeddings, cross-attention maps, and inference-time guidance. These interfaces make diffusion models especially suitable for methods such as memorization mitigation, concept unlearning, prompt mediation, safety-aligned inference, watermarking, and provenance analysis. At the same time, the dominance of diffusion-based studies should not be mistaken for architectural completeness: AR and flow-based models remain underexplored from the perspective of IP protection, and it is still unclear whether they exhibit the same vulnerabilities or require different safeguards.

Overall, the relevance of generative-model background in this survey is not merely historical. Different architectures differ in how they internalize data, expose generation behavior, and support downstream intervention. These differences directly affect both the attack surface of IP infringement and the design space of technical defenses. For this reason, the taxonomy in Section~\ref{sec:IPP} should be read not as architecture-specific, but as a control-oriented framework that must ultimately be instantiated differently across model families.

\subsection{IP Infringement}
\label{sec:back02}

Intellectual property (IP) refers to legal rights associated with creations of the mind, including inventions, literary and artistic works, designs, symbols, names, and images used in commerce; major IP regimes include copyright, trademarks, patents, and trade secrets~\cite{wipo_ip_overview,janke2018indigenous}. IP infringement, in turn, concerns the unauthorized use, reproduction, distribution, or exploitation of protected assets in ways that violate the rights granted under these regimes. In the era of visual generative AI, such infringement has become both more prevalent and more difficult to govern, largely because of three structural conditions: web-scale data collection with unclear licensing, the virtually unlimited volume of generated outputs, and the opaque and difficult-to-audit behavior of modern generative models. In this survey, we characterize these risks through two complementary asset dimensions---\textit{Data IP} and \textit{Model IP}---which align the background problem formulation with the broader taxonomy introduced in the Introduction~\ref{sec:intro}.

\textbf{Data IP.}
From the Data IP perspective, the protected assets are copyrighted or proprietary \emph{data and creative works}, including training materials, identifiable content, character images, branded assets, and expressive styles. Data IP infringement can occur both at \emph{training time} and at \emph{generation time}. At training time, the core concern is the unauthorized acquisition, copying, or use of protected material for model development. Many large-scale datasets are assembled from web-scraped content without explicit consent or clear licensing documentation, making it difficult to verify provenance or enforce rights at scale. A prominent example in visual generative AI is the Getty Images litigation against Stability AI, which alleged that Stability AI copied and processed millions of Getty-owned or Getty-represented images and associated metadata without authorization in connection with the development of Stable Diffusion~\cite{getty_stability_complaint}. Although our focus is visual generative AI, similar disputes involving books and music illustrate the broader cross-domain pattern of contested training on protected material~\cite{kadrey_meta_order,riaa_suno_complaint,riaa_udio_complaint}.

At generation time, Data IP infringement is no longer limited to verbatim copying or direct memorization. It also includes high-similarity outputs that reproduce protected expression, imitate distinctive styles, or enable derivative creation at scale. In visual generation, this may appear as style mimicry, identity replication, or near-duplicate rendering of copyrighted works. Related concerns have also been raised in code generation, where systems such as GitHub Copilot have been alleged to reproduce licensed code without proper attribution~\cite{github-copilot-investigation-url,devs-dont-rely-url}. These examples underscore that Data IP risks span the full pipeline: from the provenance and legality of training data to the similarity, traceability, and originality of generated outputs.

\textbf{Model IP.}
From the Model IP perspective, the protected assets are the \emph{models themselves}, including their weights, architectures, training recipes, alignment procedures, capability profiles, and behavior distributions. Developing competitive visual generative models requires substantial investment in computation, data curation, engineering, and deployment. As a result, the models---not only the data used to train them---must also be treated as valuable intellectual assets. Model IP infringement can therefore occur through unauthorized reuse, redistribution, fine-tuning, distillation, or functional replication of proprietary systems.

A common form of Model IP violation is the unauthorized redistribution or repackaging of model-related artifacts, such as checkpoints, adapters, or weight deltas, in ways that bypass the original model holder's intended control over access, licensing, or attribution. This risk is particularly salient for open or semi-open foundation models, where derivative checkpoints can be further modified, merged, or redistributed across downstream communities. The NovelAI case illustrates this concern in the context of derivative diffusion models: NovelAI reported an unauthorized breach involving proprietary software and source code, while its diffusion models were developed by modifying Stable Diffusion's architecture and training process~\cite{NovelAILeakAnnouncement2022,NovelAIStableDiffusionImprovements2022}. This case highlights how, once model-related artifacts or implementation assets are leaked or widely circulated, model holders may lose practical control over downstream reuse. 

Beyond artifact redistribution, model extraction and distillation-based copying pose a more direct challenge to model-level ownership. In this setting, an adversary may query a proprietary ``teacher'' model and use its outputs to train a ``student'' model that approximates the teacher's behavior, potentially becoming a functional substitute without access to the original parameters. Such risks are especially salient in API-based deployment settings, where repeated querying can reveal model behavior, response distributions, or capability signatures. Public discussion around derivative systems such as Vicuna illustrates this broader tension between reuse, derivative development, and model-level ownership claims in the era of foundation models: Vicuna was fine-tuned from LLaMA using user-shared conversations collected from ShareGPT~\cite{chiang2023vicuna}. Rather than constituting a simple case of parameter theft, such examples show how model outputs, interaction traces, and instruction-following behaviors can themselves become reusable resources for building competing or derivative systems.

Taken together, these two asset dimensions clarify why IP protection in visual generative AI cannot be reduced to a single stage of the pipeline. Some risks concern whether protected data is learned or reproduced; others concern whether proprietary model value is extracted, replicated, or repurposed. At the same time, \textit{Data IP} and \textit{Model IP} are not themselves the final taxonomy of protection methods. Rather, they define \emph{what is being protected}. In Section~\ref{sec:IPP}, we move from this asset-level framing to a control-oriented organization of defenses under the \textit{Control Logic View}, which categorizes methods by the specific risk variable they regulate: \textit{Information Exposure Control}, \textit{Generative Behavior Constraint}, and \textit{Attribution \& Accountability}.

\arrayrulecolor{black}
\definecolor{lightblue}{RGB}{220, 235, 250}  
\definecolor{lightgreen}{RGB}{220, 245, 220} 
\definecolor{lightorange}{RGB}{255, 235, 200} 

\begin{table*}[!t]
    \centering
    \renewcommand{\arraystretch}{1.4}
    \caption{IP protection methods organized by Control Logic View. \colorbox{lightblue}{Blue rows} indicate Data IP protection, and \colorbox{lightgreen}{Green rows} indicate Model IP protection.}
    \resizebox{\linewidth}{!}{
    \begin{tabular}{p{3cm}|p{10cm}|p{8cm}}
\hline 
\textbf{Subcategory} & \textbf{Protection Goal} & \textbf{Representative Methods} \\ \hline 

\rowcolor{lightorange}
\multicolumn{3}{c}{\textbf{Information Exposure Control} (Control what protected IP assets can be internalized or revealed)}\\ \hline

\rowcolor{lightblue}
Data Filtering & Remove duplicate or memorization-prone samples from training data. & 
SNIP-dedup~\cite{webster2023deduplicationlaion2b}
\\

\rowcolor{lightblue}
Data Corruption & Train models on corrupted data to prevent memorization of original samples. & 
CDMT~\cite{daras2024consistentdiffusionmeetstweedie}, 
Ambient Diffusion~\cite{daras2023ambientdiffusionlearningclean}
\\

\rowcolor{lightblue}
Compositional Training
& Train separate diffusion models on distinct data sources to enable selective forgetting.
& CDM~\cite{golatkar2024trainingdataprotectioncompositional}
\\

\rowcolor{lightblue}
Data Perturbation & Add imperceptible perturbations to images preventing unauthorized learning. & 
EUDP~\cite{zhao2024unlearnableexamplesdiffusionmodels}, 
SDS-Attack~\cite{xue2024effectiveprotectiondiffusionbased}, 
IDProtector~\cite{song2025idprotectoradversarialnoiseencoder}, 
Glaze~\cite{shan2023glazeprotectingartistsstyle},
MetaCloak~\cite{liu2024metacloak},
Nightshade~\cite{shan2024nightshade},
StyleGuard~\cite{li2025styleguard}
\\

\rowcolor{lightgreen}
Model Locking & Embed secret keys into models; unauthorized users receive degraded outputs. & 
ModelLock~\cite{gao2024modellocklockingmodelspell}, 
FMLock~\cite{liu2024fmlock},
PCDiff~\cite{gai2025pcdiff}
\\

\hline
\rowcolor{lightorange} 
\multicolumn{3}{c}{\textbf{Generative Behavior Constraint} (Control how generation behaviors may reproduce protected data or enable misuse of model capabilities)}\\ \hline

\rowcolor{lightblue}
Prompt Engineering & Detect or modify prompts to prevent IP-infringing generation requests. & 
CopyJudge~\cite{liu2025copyjudge}, 
DGP~\cite{wangevaluate}
\\

\rowcolor{lightblue}
\rowcolor{lightblue}
Concept Unlearning & Remove or suppress specific concepts (styles, objects, identities) in pretrained models. & 
ESD~\cite{gandikota2023erasing}, 
AdvUnlearn~\cite{zhang2024defensive}, 
DUO~\cite{park2024direct},
Meta-Unlearning~\cite{gao2025meta},
UCE~\cite{gandikota2024unified},
Receler~\cite{huang2024receler},
SalUn~\cite{fan2023salun},
GLoCE~\cite{lee2025localized},
NLCE~\cite{shi2026neighbor},
MACE~\cite{lu2024mace},
SAeUron~\cite{cywinski2025saeuron},
STEREO~\cite{srivatsan2025stereo},
AdaVD~\cite{wang2025precise},
Optimal-Target~\cite{bui2025fantastic}
\\

\rowcolor{lightblue}
Safety-aligned Inference & Guide inference-time sampling away from undesired or protected content. & 
SLD~\cite{schramowski2023safe}, ProtoRe~\cite{dong2023towards},
AMG~\cite{chen2024towards},
DEM~\cite{wen2024detecting}, MemAttn~\cite{ren2024unveiling}, NeMo~\cite{NEURIPS2024_a102dd59}, 
SAFREE~\cite{Yoon2025SAFREE}
\\

\rowcolor{lightgreen}
Anti-Extraction & Detect or prevent model extraction attacks via query analysis or output perturbation. & 
GAN-Defense~\cite{hu2021model}
\\

\rowcolor{lightgreen}
Personalization Defense & Prevent unauthorized fine-tuning or editing for identity/style mimicry. & 
Anti-DreamBooth~\cite{van2023anti},
PhotoGuard~\cite{salman2023raising}, 
DiffusionGuard~\cite{choi2024diffusionguard}
\\

\hline
\rowcolor{lightorange}
\multicolumn{3}{c}{\textbf{Attribution \& Accountability} (Control whether ownership or misuse can be verified, attributed, and proven)}\\ \hline

\rowcolor{lightblue}
Data Usage Verification & Detect whether specific datasets were used in model training. & Structural-MIA~\cite{li2024unveiling}, CLiD~\cite{zhai2024membership}, CDI~\cite{dubinski2024cdi}
\\

\rowcolor{lightblue}
Data Watermark & Embed identifiable watermarks into protected data for traceability. & 
ArtistsW~\cite{luo2023stealartworksfinetuningwatermarking},
GenWatermark~\cite{ma2023generative}, 
DIAGNOSIS~\cite{wang2024diagnosisdetectingunauthorizeddata}, 
FT-Shield~\cite{cui2024ftshieldwatermarkunauthorizedfinetuning}, 
DiffusionShield~\cite{cui2024diffusionshieldwatermarkcopyrightprotection}, 
EditGuard~\cite{Zhang_2024_CVPR}, 
AdvW~\cite{zhu2024watermarkembeddedadversarialexamplescopyright}, ZoDiac~\cite{zhang2024attack}
\\

\rowcolor{lightgreen}
Origin Attribution & Identify which model generated a given synthetic image. & 
RONAN~\cite{NEURIPS2023_ebb4c188}, OCC-CLIP~\cite{liu2024modelgeneratedimagemodelagnostic}, LatentTracer~\cite{wang2024tracelatentgenerativemodel}
\\

\rowcolor{lightgreen}
Model Fingerprinting & Extract unique behavioral signatures to identify model lineage or ownership. & 
GANFingerprints~\cite{yu2022artificialfingerprintinggenerativemodels},
AuthPrint~\cite{yao2025authprint},
DiffIP~\cite{li2025diffip}
\\

\rowcolor{lightgreen}
Model Watermark & Embed watermarks into models or their outputs for ownership verification and traceability. & 
WDM~\cite{zhao2023recipewatermarkingdiffusionmodels}, 
Tree-Ring~\cite{wen2023treeringwatermarksfingerprintsdiffusion}, 
StableSign~\cite{fernandez2023stablesignaturerootingwatermarks},
WaDiff~\cite{min2024watermark},
Gaussian-Shading~\cite{yang2024gaussian},
SleeperMark~\cite{wang2025sleepermarkrobustwatermarkfinetuning},
DistriMark~\cite{lei2025secureefficientwatermarkinglatent},
RivaGAN~\cite{zhang2019robustinvisiblevideowatermarking},
ROBIN~\cite{huang2024robin},
SynthID~\cite{synthid}, 
VideoShield~\cite{hu2025videoshieldregulatingdiffusionbasedvideo}, RoMa~\cite{xie2025roma}
\\

\hline
\end{tabular} }
\label{table:control_logic_view}
\end{table*}
\section{IP Protection Methods for Visual Generative Models}
\label{sec:IPP}

In this section, we systematically review IP protection methods for visual generative models under the \textit{Control Logic View} introduced above. We organize the reviewed approaches by their primary protection objective: \textit{Information Exposure Control}, \textit{Generative Behavior Constraint}, and \textit{Attribution \& Accountability}. Each objective is further analyzed across two asset dimensions, namely \textit{Data IP} protection and \textit{Model IP} protection. Table~\ref{table:control_logic_view} serves as the primary taxonomy of protection methods in this survey, while Fig.~\ref{fig:taxonomy} provides a complementary visual overview.

This taxonomy should be understood as a survey framework of \emph{intervention logic}, rather than a ranking of protection strength. \textit{Information Exposure Control} acts on what protected assets may be internalized or revealed by a model; \textit{Generative Behavior Constraint} acts on how protected content or model capabilities may be reproduced, expressed, or misused during generation; and \textit{Attribution \& Accountability} acts on whether ownership, usage, or misuse can be verified after deployment. Some methods naturally have overlapping effects across categories. In such cases, we place them according to their primary control objective, while noting cross-cutting roles when necessary. This organization highlights a key point for the survey literature: IP protection in visual generative AI is not a single technical problem, but a collection of method families that intervene at different points in the formation, expression, and verification of IP risk.

We next discuss concrete strategies for limiting information exposure, beginning with Data IP and then Model IP settings.

\subsection{Information Exposure Control}
\label{subsec:information_exposure_control}

Information Exposure Control addresses the earliest stage at which IP risks emerge: the acquisition, retention, and possible revelation of protected information by a generative model. The central question in this category is what protected assets a model can internalize during training or later reveal through memorization-like behavior. In the \textbf{Data IP} setting, the protected asset is the data itself, including copyrighted images, styles, and identity-bearing content; the main objective is therefore to reduce the model's tendency to absorb instance-specific signals that may later enable near-duplicate reproduction or leakage. In the \textbf{Model IP} setting, the protected asset is the proprietary model artifact and its commercial capability; the objective is to prevent unauthorized users from obtaining useful value from the model, even if they gain access to checkpoints or deployment interfaces. Compared with later categories, exposure control is the most explicitly preventive branch of the taxonomy, because it attempts to reduce risk before infringing behavior or ownership disputes materialize.

\subsubsection{Data IP Protection}
\label{subsubsec:exposure_data_ip}

Data-side exposure control aims to make protected information less likely to be memorized or learnable in the first place. Existing methods can be broadly grouped into four families: \emph{data filtering}, \emph{data corruption}, \emph{compositional training}, and \emph{data perturbation}. Although these families differ technically, they share the common goal: weakening the pathway from protected training data to later memorization or regurgitation.

\paragraph{Data Filtering}
Data filtering reduces exposure at the source by removing duplicate or memorization-prone samples before training. The basic intuition is that repeated exposure to highly similar samples disproportionately increases the likelihood of instance-level recall. \textit{SNIP-dedup}~\cite{webster2023deduplicationlaion2b} is a representative example, performing large-scale deduplication to identify and remove duplicate or near-duplicate samples from web-scale corpora. In the existing literature, filtering-based methods are attractive because they are conceptually simple and model-agnostic, but they depend heavily on accurate similarity search and cannot address memorization arising from unique yet still highly sensitive samples.

\paragraph{Data Corruption}
Instead of removing data, corruption-based methods reduce exposure by degrading the fidelity of the observed training signal, thereby encouraging the model to learn distribution-level regularities rather than precise instance-specific details. \textit{CDMT}~\cite{daras2024consistentdiffusionmeetstweedie} trains diffusion models on corrupted observations so that the learned denoising process becomes less reliant on exact original samples. \textit{Ambient Diffusion}~\cite{daras2023ambientdiffusionlearningclean} generalizes this idea by treating training data as outputs of a measurement process and training the model under such measurement operators. Corruption-based methods occupy an intermediate position between filtering and full data transformation: unlike filtering, they retain all samples, but unlike perturbation-based protection, they require control over the training pipeline itself. Their main strength is that they reduce memorization pressure without necessarily discarding large portions of the dataset.

\paragraph{Compositional Training}
Compositional training addresses exposure structurally by partitioning protected assets across multiple components, so that no single model fully internalizes all sensitive sources. \textit{CDM}~\cite{golatkar2024trainingdataprotectioncompositional} exemplifies this design by training separate diffusion components on distinct data sources and enabling controlled composition at inference time. The significance of this family is that it reframes IP protection as an architectural modularization problem: exposure is reduced not only by modifying data, but by changing how knowledge is allocated across model components. This also makes selective source removal more feasible, since protected components can be disabled or replaced without retraining the entire system.

\paragraph{Data Perturbation}
Data perturbation is one of the most proactive data-owner-side defenses, because it remains applicable even when downstream training is completely outside the owner's control. The common strategy is to embed imperceptible perturbations that preserve human perceptual quality while making the data harder to learn or causing systematic failure under unauthorized training. Within this family, existing methods differ by what aspect of learning they disrupt. \textit{EUDP}~\cite{zhao2024unlearnableexamplesdiffusionmodels} makes samples unlearnable for diffusion training; \textit{SDS-Attack}~\cite{xue2024effectiveprotectiondiffusionbased} interferes more directly with diffusion learning signals through score-distillation-style optimization; \textit{IDProtector}~\cite{song2025idprotectoradversarialnoiseencoder} specifically targets identity learning in subject-driven personalization; and \textit{Glaze}~\cite{shan2023glazeprotectingartistsstyle} focuses on style cloaking for artist protection. More recent work strengthens either transferability or robustness: \textit{MetaCloak}~\cite{liu2024metacloak} improves cross-model transfer under a meta-learning objective, \textit{Nightshade}~\cite{shan2024nightshade} adopts a stronger poisoning-style strategy that induces harmful concept associations in unauthorized training, and \textit{StyleGuard}~\cite{li2025styleguard} emphasizes practicality and robustness against style mimicry. 

Taken together, these four families reveal a common theme in data-side exposure control: the primary aim is not yet to constrain outputs, but to reduce the probability that protected content becomes stably encoded and recoverable in the learned model at all. Filtering and corruption assume access to the training process, whereas perturbation empowers data owners directly; compositional training, by contrast, changes the structural unit of exposure itself.

\subsubsection{Model IP Protection}
\label{subsubsec:exposure_model_ip}

In the \textbf{Model IP} setting, Information Exposure Control is centered on preventing unauthorized users from extracting useful value from a proprietary model. Compared with data-side exposure control, the literature here is narrower but conceptually clear: the dominant family is \emph{model locking}, which binds usable behavior to authorization.

\paragraph{Model Locking}
Model locking ties normal generation quality to a secret credential or unlocking condition, so that stolen checkpoints or unauthorized deployments yield degraded behavior. \textit{ModelLock}~\cite{gao2024modellocklockingmodelspell} is a representative example, embedding a secret unlocking condition such that high-quality outputs require the correct key or prompt condition, while unauthorized usage produces degraded generations. \textit{FMLock}~\cite{liu2024fmlock} similarly binds usability to a secret key, but implements this dependence through a key-driven mechanism that changes model behavior according to whether the correct key is present. \textit{PCDiff}~\cite{gai2025pcdiff} extends this direction to diffusion-specific proactive protection while also considering compatibility with watermarking.

Model locking is important because it captures a distinctly model-centric version of exposure control: the concern is not whether the model memorizes external data, but whether unauthorized parties can benefit from proprietary capability after obtaining the model. This family is still relatively small compared with watermarking or attribution, which also highlights one of the broader gaps in the literature: proactive model-level safeguards remain much less developed than reactive ownership verification.

\subsection{Generative Behavior Constraint}
\label{subsec:generative_behavior_constraint}

Generative Behavior Constraint assumes that some protected information or proprietary capability may already exist in a trained model, and focuses on limiting how it can be expressed during generation. In other words, the target is no longer what the model has learned, but what it is allowed to do with that information at inference time. For \textbf{Data IP}, this means reducing the generation of infringing, high-similarity, or style-mimicking outputs. For \textbf{Model IP}, it means limiting the transfer, cloning, or misuse of proprietary capability through querying, editing, personalization, or distillation. This is one of the most practically important branches of the taxonomy, because many deployed systems cannot be retrained from scratch and instead require downstream control at the level of prompts, latent states, attention, or outputs.

\subsubsection{Data IP Protection}
\label{subsubsec:behavior_data_ip}

From the Data IP perspective, Generative Behavior Constraint encompasses methods that suppress infringing outputs while preserving utility on unrelated content. In the current literature, three major families dominate: \emph{prompt engineering}, \emph{concept unlearning}, and \emph{safety-aligned inference}. These families differ mainly in \emph{where} they intervene: before generation begins, through model-level capability editing, or during the sampling process itself.

\paragraph{Prompt Engineering}
Prompt-level methods attempt to prevent infringement by identifying risky requests or modifying conditioning signals before generation proceeds. Their main advantage is deployment convenience: they can often be integrated into existing systems without updating the base model. \textit{CopyJudge}~\cite{liu2025copyjudge} exemplifies this line through a detection-and-mitigation pipeline that flags potentially infringing prompts and supports actions such as rewriting or refusal. \textit{DGP}~\cite{wangevaluate} similarly performs prompt-conditioned guidance to reduce the likelihood that diffusion trajectories converge to protected content.

At a family level, prompt engineering is best viewed as a upstream behavioral filter. It is lightweight and practical for commercial systems, but its protection power is often limited by prompt ambiguity, jailbreak-like rewriting, and the fact that the underlying capability remains intact.

\paragraph{Concept Unlearning}
Concept unlearning is one of the most heavily studied families in visual IP protection because it directly addresses whether a pretrained model can be prevented from generating a protected concept, such as a style, identity, object, logo, or branded content. Prior work can be understood along several design axes rather than as an undifferentiated enumeration of methods.

A first group consists of \emph{parameter-update-based erasure} methods, which directly modify model weights to suppress the target concept. \textit{ESD}~\cite{gandikota2023erasing} is the canonical starting point in this line, fine-tuning model parameters toward forgetting. \textit{AdvUnlearn}~\cite{zhang2024defensive} strengthens this design by explicitly balancing forgetting and retention to reduce collateral damage. \textit{DUO}~\cite{park2024direct} similarly optimizes direct forgetting objectives, while \textit{Meta-Unlearning}~\cite{gao2025meta} addresses persistence by making erased concepts harder to recover through subsequent adaptation. These methods are appealing because they target the capability itself, but they must balance forgetting effectiveness against side effects on unrelated generation quality.

A second group focuses on \emph{efficient, training-free, and localized editing}. \textit{UCE}~\cite{gandikota2024unified} provides a training-free concept editing approach, reducing the need for expensive retraining. \textit{Receler}~\cite{huang2024receler} emphasizes lightweight erasure interventions, and \textit{SalUn}~\cite{fan2023salun} updates only salient parameters to achieve more parameter-efficient forgetting. More recently, localized concept erasure has emerged as an important direction within this group. \textit{GLoCE}~\cite{lee2025localized} introduces a training-free gated low-rank adaptation framework that selectively removes target concepts only in the spatial regions where they appear, thereby better preserving non-target regions and overall image fidelity. Building on this localized perspective, \textit{NLCE}~\cite{shi2026neighbor} further incorporates neighbor-aware preservation, aiming to erase the target concept while maintaining semantically related neighboring concepts, which is particularly important in fine-grained settings. These methods reflect a broader trend in the literature toward practical deployment, where the goal is not only to erase concepts efficiently, but also to improve locality, specificity, and preservation of non-target content.

A third group studies \emph{scalability, interpretability, and robustness}. \textit{MACE}~\cite{lu2024mace} investigates scalable multi-concept erasure while preserving overall generation quality. \textit{SAeUron}~\cite{cywinski2025saeuron} moves the intervention from parameter space to sparse feature space, improving interpretability and target specificity. \textit{STEREO}~\cite{srivatsan2025stereo} explicitly strengthens robustness against adversarial prompt-based bypass, \textit{AdaVD}~\cite{wang2025precise} uses orthogonalization to better separate target and retained directions, and \textit{Optimal-Target}~\cite{bui2025fantastic} asks a more fundamental question: what is the most effective erasure target to optimize against in the first place.

Taken together, concept unlearning should be understood not as a single method, but as a rich design space spanning weight-level forgetting, training-free editing, localized concept erasure, parameter-efficient intervention, interpretable feature suppression, preservation of semantic neighbors, and robustness to relearning or bypass. Its main strength is that it directly weakens the offending generation capability; its main challenge is how to do so precisely, robustly, and without unacceptable side effects.

\paragraph{Safety-aligned Inference}
Safety-aligned inference methods constrain generation during sampling rather than through permanent weight changes. They are therefore particularly relevant when retraining is infeasible or when practitioners require tunable deployment-time control. Existing work in this family differs mainly in what part of the inference trajectory is manipulated. \textit{SLD}~\cite{schramowski2023safe} introduces safety-oriented guidance that biases diffusion trajectories away from undesirable regions. \textit{ProtoRe}~\cite{dong2023towards} performs prototype-based concept negation at inference time. \textit{AMG}~\cite{chen2024towards} steers generation away from memorized instances while maintaining text-image alignment. \textit{DEM}~\cite{wen2024detecting} combines detection and mitigation during sampling. \textit{MemAttn}~\cite{ren2024unveiling} alters attention behavior to reduce memorization-related recall, \textit{NeMo}~\cite{NEURIPS2024_a102dd59} suppresses memorization-associated neurons, and \textit{SAFREE}~\cite{Yoon2025SAFREE} performs adaptive subspace steering in the conditioning space.

Compared with concept unlearning, inference-time methods are usually more deployment-friendly and reversible, but they often rely more heavily on the quality of runtime guidance and may be more vulnerable to adaptive prompting or other bypass strategies. They are best understood as flexible behavioral overlays rather than true removal of underlying capability.

\subsubsection{Model IP Protection}
\label{subsubsec:behavior_model_ip}

In the \textbf{Model IP} setting, Generative Behavior Constraint focuses on preventing proprietary capability from being cloned, transferred, or repurposed through downstream interaction. The two main families are \emph{anti-extraction} and \emph{personalization defense}. The latter is inherently cross-cutting, since unauthorized personalization may simultaneously harm Data IP (e.g., identity or style misuse) and Model IP (e.g., capability transfer). In this survey, we place it under Model IP because its primary threat arises from the unauthorized reuse of generative capability.

\paragraph{Anti-Extraction}
Anti-extraction methods aim to make it harder for attackers to clone a proprietary generator through repeated querying, imitation, or distillation. \textit{GAN-Defense}~\cite{hu2021model} is an early representative, perturbing outputs or altering query-response behavior so that a black-box attacker cannot easily train a high-fidelity substitute model. Although the literature here is still relatively sparse compared with data-side protection, it is conceptually important because it frames Model IP risk as a behavioral cloning problem rather than merely a checkpoint theft problem.

More broadly, this family reflects a shift in IP thinking for AI systems: even without direct parameter access, a model may still be economically replicated through sufficiently informative interaction. That makes anti-extraction a core component of Model IP protection under API-based deployment.

\paragraph{Personalization Defense}
Personalization defenses aim to prevent unauthorized fine-tuning, subject-driven generation, or editing pipelines from turning a general-purpose model into a customized infringing tool. \textit{Anti-DreamBooth}~\cite{van2023anti} perturbs user-shared images so that DreamBooth-style personalization yields poor identity fidelity. \textit{PhotoGuard}~\cite{salman2023raising} immunizes images against malicious diffusion-based editing by introducing subtle protective perturbations. \textit{DiffusionGuard}~\cite{choi2024diffusionguard} further strengthens this line by targeting more adaptive malicious editing and personalization settings.

The significance of this family is that it broadens the notion of Model IP misuse: the risk is not only that a model is copied, but also that it is adapted into a high-performing derivative service specialized for imitation, impersonation, or targeted style replication. This makes personalization defense a natural bridge between model-level capability protection and data-side identity/style protection.

\subsection{Attribution \& Accountability}
\label{subsec:attribution_accountability}

Attribution \& Accountability provides a more reactive line of defense. Rather than preventing infringement from occurring, methods in this category enable auditing, tracing, ownership verification, and evidence collection after a dispute has arisen. In the \textbf{Data IP} setting, the central question is whether protected data influenced a deployed model or output. In the \textbf{Model IP} setting, the question becomes whether a model, its outputs, or its behavior can be attributed to a legitimate source or owner. This branch is especially important in realistic governance settings, where preventive defenses are incomplete and enforceability often depends on post-hoc evidence.

\subsubsection{Data IP Protection}
\label{subsubsec:aa_data_ip}

From the Data IP perspective, Attribution \& Accountability includes both \emph{data usage verification} and \emph{data watermarking}. The former infers whether a protected dataset was used during training; the latter embeds traceable signals so that later use leaves direct evidence.

\paragraph{Data Usage Verification}
These methods aim to test whether a model was trained on a specific dataset or data source, thereby enabling data owners to substantiate usage claims. \textit{Structural-MIA}~\cite{li2024unveiling} represents membership-style verification approaches that infer dataset usage from model behavior or structural signals. \textit{CLiD}~\cite{zhai2024membership} provides diffusion-specific identification signals for dataset-level usage testing, while \textit{CDI}~\cite{dubinski2024cdi} offers a more explicit verification pipeline based on model response characteristics.

These methods are important because they provide \emph{forensic evidence without proactive intervention}. Their main limitation, however, is that they often depend on statistical inference quality and may be harder to interpret or defend legally than explicit embedded signals.

\paragraph{Data Watermark}
Data watermarking embeds identifiable or traceable signals into protected data, or related artifacts, such that downstream unauthorized use leaves detectable evidence. Different methods in this family vary mainly in where the evidence is intended to surface: in training effects, in generated outputs, or under adversarial post-processing. \textit{ArtistsW}~\cite{luo2023stealartworksfinetuningwatermarking} emphasizes artist-centric traceability; \textit{GenWatermark}~\cite{ma2023generative} studies watermark signals recoverable from generated images; \textit{DIAGNOSIS}~\cite{wang2024diagnosisdetectingunauthorizeddata} explicitly embeds signals for detecting unauthorized dataset usage; and \textit{FT-Shield}~\cite{cui2024ftshieldwatermarkunauthorizedfinetuning} together with \textit{DiffusionShield}~\cite{cui2024diffusionshieldwatermarkcopyrightprotection} tailor this approach to diffusion-model settings. More robust variants such as \textit{EditGuard}~\cite{Zhang_2024_CVPR}, \textit{AdvW}~\cite{zhu2024watermarkembeddedadversarialexamplescopyright}, and \textit{ZoDiac}~\cite{zhang2024attack} further strengthen traceability under editing, adversarial removal, or transformation.

Compared with usage verification, data watermarking is more proactive because it embeds evidence before infringement occurs. Its main challenge is the familiar watermarking tradeoff among invisibility, detectability, robustness, and transferability through training and generation.

\subsubsection{Model IP Protection}
\label{subsubsec:aa_model_ip}

In the \textbf{Model IP} setting, Attribution \& Accountability methods aim to identify the source of a generated artifact, verify ownership of a deployed model, or preserve provenance evidence under redistribution, fine-tuning, and laundering. The literature here is commonly grouped into \emph{origin attribution}, \emph{fingerprinting}, and \emph{model watermarking}. These families are related, but they differ in what counts as evidence: inferred source identity, elicited behavioral signature, or explicitly embedded mark.

\paragraph{Origin Attribution}
Origin attribution asks a direct question: \emph{which model generated this output?} \textit{RONAN}~\cite{NEURIPS2023_ebb4c188} approaches this problem through inversion- or reconstruction-based evidence that differentiates likely source models. \textit{OCC-CLIP}~\cite{liu2024modelgeneratedimagemodelagnostic} formulates source identification as a classifier-style attribution problem under limited supervision. \textit{LatentTracer}~\cite{wang2024tracelatentgenerativemodel} leverages latent inversion and reconstruction consistency to trace images back to the most plausible generating model.

Origin attribution is best understood as the least assumption-heavy evidence family in this branch: it does not necessarily require proactive watermark insertion, but instead exploits post-hoc distinguishability among candidate generators.

\paragraph{Model Fingerprinting}
Fingerprinting methods identify model lineage or ownership through stable behavioral signatures. Unlike watermarking, which often relies on deliberately embedded identifiers, fingerprinting may exploit intrinsic or elicited patterns in model behavior. \textit{GANFingerprints}~\cite{yu2022artificialfingerprintinggenerativemodels} introduces detectable fingerprint signals from generated outputs; \textit{AuthPrint}~\cite{yao2025authprint} verifies ownership by eliciting model-specific responses to designed queries; and \textit{DiffIP}~\cite{li2025diffip} adapts this logic to diffusion models for robust black-box or gray-box identity verification.

In the literature, a useful distinction is that fingerprinting sits between attribution and ownership proof. It is not just about identifying a likely source model, but about demonstrating that a model exhibits a stable and distinguishable behavioral identity.

\paragraph{Model Watermark}
Model watermarking embeds verifiable signals into the model or its generation process so that ownership can still be established after redistribution, fine-tuning, or laundering attempts. Because this is one of the largest literature families in Model IP protection, it is useful to understand it along several design axes.

A first group embeds signatures more directly into the model or generation behavior. \textit{WDM}~\cite{zhao2023recipewatermarkingdiffusionmodels} provides a watermarking mechanism for diffusion models with explicit detectability objectives, while \textit{StableSign}~\cite{fernandez2023stablesignaturerootingwatermarks} introduces a lightweight signature tied to generation behavior. \textit{SleeperMark}~\cite{wang2025sleepermarkrobustwatermarkfinetuning} explicitly tackles a central challenge in this literature, namely watermark survival after subsequent fine-tuning.

A second group leverages \emph{noise- or latent-space embedding}, which is particularly natural for diffusion models. \textit{Tree-Ring}~\cite{wen2023treeringwatermarksfingerprintsdiffusion} encodes structured signals in the initialization or noise domain, enabling robust watermark recovery under common transformations. \textit{DistriMark}~\cite{lei2025secureefficientwatermarkinglatent} emphasizes distribution-level detectability and robustness under diverse outputs and attacker perturbations. \textit{WaDiff}~\cite{min2024watermark} and \textit{Gaussian-Shading}~\cite{yang2024gaussian} explore related diffusion-oriented watermarking strategies with different robustness-design tradeoffs.

A third group emphasizes \emph{deployment practicality and media coverage}. \textit{RivaGAN}~\cite{zhang2019robustinvisiblevideowatermarking} provides a robust watermarking backbone for generated media, \textit{ROBIN}~\cite{huang2024robin} emphasizes practical robustness, \textit{SynthID}~\cite{synthid} targets large-scale standardized watermarking for synthetic media, \textit{VideoShield}~\cite{hu2025videoshieldregulatingdiffusionbasedvideo} extends watermarking to diffusion-based video with spatiotemporal robustness, and \textit{RoMa}~\cite{xie2025roma} contributes another robust watermarking mechanism for diffusion-based content.

Overall, model watermarking is the most explicit evidence-embedding family in the Model IP branch. Its core challenge is not only detectability, but persistence under realistic post-processing, fine-tuning, laundering, and distribution shifts. This makes it one of the clearest examples of how technical design and legal evidentiary needs intersect in IP protection research.

In summary, the literature reviewed in this section shows that visual IP protection is best understood as a collection of intervention families that target different risk variables rather than a single unified defense recipe. Some methods act before protected information is absorbed, others constrain how it is expressed during generation, and still others provide evidence once infringement has occurred. This control-oriented organization clarifies common design patterns across otherwise heterogeneous techniques, while also revealing structural gaps in the field, such as the relative scarcity of proactive model-level defenses and the limited cross-architecture understanding beyond diffusion models. In the next section, we show that this same taxonomy can also organize evaluation, allowing protection methods and their assessment to be discussed within a unified analytical framework.

\section{Evaluation of IP Protection Methods}
\label{sec:eval}

\arrayrulecolor{black}
\definecolor{lightblue}{RGB}{220, 235, 250}  
\definecolor{lightgreen}{RGB}{220, 245, 220} 
\definecolor{lightorange}{RGB}{255, 235, 205} 

\begin{table*}[!t]
    \centering
    \renewcommand{\arraystretch}{1.25}
    \caption{Evaluation metrics for IP protection organized by Control Logic View. \colorbox{lightblue}{Blue rows} indicate Data IP metrics, and \colorbox{lightgreen}{Green rows} indicate Model IP metrics.}
    \label{table:metric}
    \resizebox{\linewidth}{!}{%
    \begin{tabular}{C{1.55cm}|C{1.65cm}|L{7.4cm}|L{6.6cm}}
    \hline
    \textbf{Category} & \textbf{Dimension} & \multicolumn{1}{c|}{\textbf{Description}} & \multicolumn{1}{c}{\textbf{Metrics}} \\ 
    \hline

    \rowcolor{lightorange}
    \multicolumn{4}{c}{\textbf{Information Exposure Control}} \\ 
    \hline

    \rowcolor{lightblue}
    \multirow{2}{*}{\makecell[c]{Data IP}} 
    & \makecell[c]{Effectiveness}
    & Measures whether the protection prevents data memorization, duplication, or unauthorized learning from protected samples.
    & L2-Norm~\cite{carlini2023extracting}; Deduplication Rate~\cite{webster2023deduplicationlaion2b}; SSCD Score~\cite{pizzi2022selfsuperviseddescriptorimagecopy}; PSNR/SSIM/LPIPS~\cite{zhang2018unreasonable,cao2023impress}; FID~\cite{heusel2017gans} \\
    \cline{2-4}

    \rowcolor{lightblue}
    & \makecell[c]{Robustness}
    & Evaluates whether data-side protection (e.g., perturbations, poisons) remains effective under purification or preprocessing attacks.
    & Protection Rate under DiffPure~\cite{nie2022diffusion}; Protection Rate under JPEG/Gaussian Noise~\cite{shan2023glazeprotectingartistsstyle,liu2024metacloak} \\
    \hline

    \rowcolor{lightgreen}
    \multirow{2}{*}{\makecell[c]{Model IP}} 
    & \makecell[c]{Effectiveness}
    & Measures whether locked models correctly authenticate authorized users and degrade outputs for unauthorized access.
    & Locked/Unlocked Downstream Performance~\cite{gao2024modellocklockingmodelspell}; FID Gap~\cite{liu2024fmlock} \\
    \cline{2-4}

    \rowcolor{lightgreen}
    & \makecell[c]{Robustness}
    & Evaluates resistance of model locking mechanisms to key recovery or fine-tuning bypass.
    & Unlocking Attack Success Rate~\cite{gao2024modellocklockingmodelspell}; Key Recovery Rate~\cite{liu2024fmlock}; Fine-tuning Bypass Rate~\cite{liu2024fmlock} \\
    \hline

    \rowcolor{lightorange}
    \multicolumn{4}{c}{\textbf{Generative Behavior Constraint}} \\ 
    \hline

    \rowcolor{lightblue}
    \multirow{2}{*}{\makecell[c]{Data IP}} 
    & \makecell[c]{Effectiveness}
    & Measures how effectively the method removes target concepts or suppresses copyrighted content during generation.
    & Detection Accuracy/F1~\cite{liu2025copyjudge}; Unlearning Accuracy~\cite{zhang2024unlearncanvasstylizedimagedataset,cheng2024mu}; Remaining Accuracy~\cite{zhang2024unlearncanvasstylizedimagedataset}/FID/CLIP Score~\cite{heusel2017gans,hessel2021clipscore}; Memorization Score~\cite{wen2024detecting,ren2024unveiling} \\
    \cline{2-4}

    \rowcolor{lightblue}
    & \makecell[c]{Robustness}
    & Evaluates whether constrained behaviors can be bypassed via adversarial prompts or concept inversion attacks.
    & Attack Success or Concept Recovery Rate~\cite{tsai2023ring,pham2023circumventing,zhang2024defensive,gong2024reliable,zhang2024generate} \\
    \hline

    \rowcolor{lightgreen}
    \multirow{2}{*}{\makecell[c]{Model IP}} 
    & \makecell[c]{Effectiveness}
    & Measures the ability to detect or prevent model extraction and block unauthorized personalization.
    & Attack Model Fidelity~\cite{hu2021model}; Identity Score Matching~\cite{van2023anti}; CLIP similarity~\cite{radford2021learning}; ImageReward~\cite{xu2023imagereward} \\
    \cline{2-4}

    \rowcolor{lightgreen}
    & \makecell[c]{Robustness}
    & Evaluates protection persistence against different personalization methods or adaptive extraction attacks.
    & Defense Rate under DreamBooth~\cite{ruiz2023dreambooth}/Textual Inversion~\cite{gal2022image}; Adaptive Attack Success Rate~\cite{van2023anti,choi2024diffusionguard} \\
    \hline

    \rowcolor{lightorange}
    \multicolumn{4}{c}{\textbf{Attribution \& Accountability}} \\ 
    \hline

    \rowcolor{lightblue}
    \multirow{2}{*}{\makecell[c]{Data IP}} 
    & \makecell[c]{Effectiveness}
    & Measures accuracy of detecting unauthorized data usage or verifying embedded data watermarks.
    & TPR at fixed FPR/AUC~\cite{zhai2024membership}; Detection Rate/Bit Accuracy~\cite{cui2024ftshieldwatermarkunauthorizedfinetuning,wang2024diagnosisdetectingunauthorizeddata,zhang2024attack}; PSNR/SSIM/LPIPS~\cite{zhang2018unreasonable,zhang2024attack} \\
    \cline{2-4}

    \rowcolor{lightblue}
    & \makecell[c]{Robustness}
    & Evaluates detection or watermark consistency under image-level transformations or watermark removal attacks.
    & Detection Rate/Bit Accuracy under Corruptions~\cite{cui2024ftshieldwatermarkunauthorizedfinetuning,Zhang_2024_CVPR}; Removal Attack Success Rate~\cite{zhang2024attack} \\
    \hline

    \rowcolor{lightgreen}
    \multirow{2}{*}{\makecell[c]{Model IP}} 
    & \makecell[c]{Effectiveness}
    & Measures accuracy of attributing generated images to source models, extracting model fingerprints, or detecting embedded model watermarks.
    & Attribution Accuracy~\cite{NEURIPS2023_ebb4c188,wang2024tracelatentgenerativemodel}; Bit Accuracy/TPR at fixed FPR/$p$-value~\cite{yu2022artificialfingerprintinggenerativemodels,wen2023treeringwatermarksfingerprintsdiffusion,zhao2023recipewatermarkingdiffusionmodels,fernandez2023stablesignaturerootingwatermarks,li2025diffip}; FID~\cite{heusel2017gans} \\
    \cline{2-4}

    \rowcolor{lightgreen}
    & \makecell[c]{Robustness}
    & Evaluates fingerprint or watermark persistence under image-level or model-level perturbations.
    & Bit-Acc under Image Attacks~\cite{jiang2023evading,fernandez2023stablesignaturerootingwatermarks,zhao2023recipewatermarkingdiffusionmodels,huang2024robin,lei2025secureefficientwatermarkinglatent}; Model Fine-tuning/Pruning~\cite{zhao2023recipewatermarkingdiffusionmodels,wang2025sleepermarkrobustwatermarkfinetuning,li2025diffip} \\
    \hline

    \end{tabular}%
    }
\end{table*}

Evaluation is essential for turning IP protection from a collection of defensive ideas into a comparable and testable research field. In visual generative AI, evaluation is intrinsically heterogeneous: different protection methods are designed to regulate different IP risk variables and therefore cannot be assessed by a single universal metric. For example, a method that reduces memorization should not be evaluated in the same way as a watermarking method for ownership proof, nor should a model-locking mechanism be judged by the same criteria as concept unlearning. Evaluation in IP protection must therefore be \emph{goal-aware} and aligned with the underlying protection objective.

Following the \textit{Control Logic View}, we organize evaluation protocols according to the same three control objectives used to review protection methods: \textit{Information Exposure Control}, \textit{Generative Behavior Constraint}, and \textit{Attribution \& Accountability}. Within each objective, we further distinguish between \textit{Data IP} and \textit{Model IP} evaluation. Across these settings, two dimensions are central: \textit{Effectiveness}, which measures whether a method achieves its intended protection goal, and \textit{Robustness}, which measures whether the protection persists under attacks, transformations, removals, or adaptive misuse. As summarized in Table~\ref{table:metric}, this organization provides a unified evaluation map across control objectives, asset dimensions, and evaluation dimensions.

It is also important to distinguish IP-specific evaluation from general-purpose quality assessment. Metrics such as FID, CLIP-based alignment scores, perceptual similarity, runtime, and memory overhead may appear across multiple categories. These metrics are useful, but they are usually \emph{auxiliary} rather than primary: they assess whether protection comes at a significant cost to visual quality, alignment, or efficiency. The main evaluation question is therefore not only whether a method preserves utility, but whether it succeeds on the specific IP risk it is intended to control.

\subsection{Evaluation for Information Exposure Control}
\label{sec:eval_info}

Information Exposure Control methods aim to prevent protected assets from being internalized or revealed by the model. Accordingly, evaluation in this category focuses on whether protected information is less learnable, less memorized, or less usable after protection. In the \textbf{Data IP} setting, this mainly concerns memorization and unauthorized learning from sensitive training samples. In the \textbf{Model IP} setting, the concern shifts to whether unauthorized users can still access proprietary model value despite protection. 

\subsubsection{Data IP Evaluation}
\label{sec:eval_info_data}

Data IP evaluation in this category assesses whether protection methods can effectively reduce memorization, near-duplicate reproduction, or unauthorized learning from protected samples. The central challenge is that memorization is not directly observable; it must be inferred through similarity, retrieval, or leakage-oriented proxies.

\paragraph{Effectiveness}Effectiveness is primarily evaluated by quantifying how closely generated samples resemble protected training instances, and by testing whether protective preprocessing successfully reduces such resemblance.

A first family of metrics measures \emph{instance-level or feature-level similarity} between generated outputs and training samples. \textbf{L2-Norm Distance}~\cite{carlini2023extracting} measures pixel-level similarity to the nearest training neighbor, with lower values indicating a higher memorization risk. \textbf{SSCD Score}~\cite{pizzi2022selfsuperviseddescriptorimagecopy} measures object-level similarity using Self-Supervised Copy Detection embeddings and is especially useful when pixel-level comparison is too brittle to capture semantic duplication.

A second family of metrics measures \emph{dataset redundancy or protection outcome}. \textbf{Deduplication Rate}~\cite{webster2023deduplicationlaion2b} quantifies the proportion of near-duplicate samples identified and removed from the dataset, and is therefore especially relevant for data filtering methods. Although not a memorization metric by itself, it provides a direct estimate of how strongly repeated exposure has been reduced.

A third family of metrics captures \emph{perceptual similarity and generation quality}. \textbf{PSNR}, \textbf{SSIM}, and \textbf{LPIPS}~\cite{zhang2018unreasonable,cao2023impress} are widely used to assess structural or perceptual similarity between images, especially in perturbation-based settings where one must verify that protection does not excessively distort the original sample. \textbf{FID}~\cite{heusel2017gans} is often used as an auxiliary metric to assess whether exposure-control methods maintain acceptable generation quality and diversity while reducing memorization.

Taken together, these metrics reflect a broader principle: Data IP exposure control is effective only if it weakens memorization-related similarity while preserving acceptable training or generation utility.

\paragraph{Robustness}
Robustness in this category evaluates whether data-side protection remains effective after purification, preprocessing, or other transformations designed to neutralize the protection signal. This is especially important for perturbation-based methods, since an attacker may attempt to remove or weaken the perturbation before training.

Common robustness evaluations are therefore attack-conditioned \emph{protection rates}. \textbf{Protection Rate under DiffPure}~\cite{nie2022diffusion,shan2023glazeprotectingartistsstyle,liu2024metacloak} measures whether protective perturbations survive diffusion-based purification. \textbf{Protection Rate under JPEG Compression}~\cite{shan2023glazeprotectingartistsstyle,liu2024metacloak} evaluates resistance to lossy compression, while \textbf{Protection Rate under Gaussian Noise}~\cite{shan2023glazeprotectingartistsstyle,liu2024metacloak} tests resilience to noise-based preprocessing. These evaluations reflect a key design requirement of proactive data protection: protection signals must remain effective under realistic transformations that an unauthorized trainer may apply before model training.

\subsubsection{Model IP Evaluation}
\label{sec:eval_info_model}

Model IP evaluation in this category focuses primarily on model locking and related access-control mechanisms. The core question is whether protected models remain useful for authorized users while becoming significantly less useful to unauthorized ones.

\paragraph{Effectiveness}
The effectiveness of model locking is typically measured by contrasting model behavior under authorized and unauthorized access conditions. A first commonly used metric is \textbf{Locked/Unlocked Downstream Performance}~\cite{gao2024modellocklockingmodelspell}, which compares task performance or generation quality when the correct key is present versus absent. A large gap indicates that authorization meaningfully controls model usability.

A second metric is the \textbf{FID Gap}~\cite{liu2024fmlock}, which specifically measures the difference in generation quality between authorized and unauthorized usage. This is particularly useful in generative settings, where downstream task accuracy alone may not fully reflect degradation in output fidelity. Overall, these metrics capture the central requirement of model-side exposure control: normal-quality functionality should remain available to legitimate users while being substantially degraded for illegitimate ones.

\paragraph{Robustness}
Robustness in model locking focuses on whether adversaries can bypass, infer, or recover the authorization mechanism. \textbf{Unlocking Attack Success Rate}~\cite{gao2024modellocklockingmodelspell} measures how often adversarial attempts succeed in using the model without the correct key. \textbf{Key Recovery Rate}~\cite{liu2024fmlock} evaluates whether attackers can reconstruct or infer the secret credential itself. \textbf{Fine-tuning Bypass Rate}~\cite{liu2024fmlock} tests whether unauthorized users can recover acceptable model performance simply by fine-tuning the protected model.

These robustness metrics are especially important because model locking is only meaningful if the gap between authorized and unauthorized usage cannot be easily erased by reverse engineering or post hoc adaptation.

\subsection{Evaluation for Generative Behavior Constraint}
\label{sec:eval_behavior}

Generative Behavior Constraint methods regulate how protected content or capabilities may be expressed during generation. Evaluation in this category therefore focuses on two coupled questions: whether protected behavior is successfully suppressed, and whether non-protected behavior is preserved. In practice, this category almost always requires a balance between \emph{target suppression} and \emph{utility retention}, together with robustness against adversarial bypass.

\subsubsection{Data IP Evaluation}
\label{sec:eval_behavior_data}

In the \textbf{Data IP} setting, evaluation assesses whether protected concepts, identities, styles, or copyrighted content are successfully suppressed during generation, while utility on unrelated prompts remains intact.

\paragraph{Effectiveness}
A first family of metrics evaluates \emph{target suppression}. \textbf{Detection Accuracy/F1}~\cite{liu2025copyjudge} is relevant for prompt-engineering systems that first identify whether a request or output is potentially infringing. In concept-unlearning settings, the most widely used metric is \textbf{Unlearning Accuracy (UA)}~\cite{zhang2024unlearncanvasstylizedimagedataset, cheng2024mu}, which measures how often the model fails to produce the erased target concept when prompted accordingly. A higher UA indicates more successful forgetting or suppression. These accuracy-based metrics are typically computed using pretrained evaluation models or classifiers.

A second family evaluates \emph{utility retention}. \textbf{Remaining Accuracy (RA)}~\cite{zhang2024unlearncanvasstylizedimagedataset} measures how well the model continues to generate correct outputs for concepts unrelated to the target of removal. This is especially important in unlearning and editing settings, where side effects on benign concepts are often the major practical concern. More generally, \textbf{FID} and \textbf{CLIP Score}~\cite{heusel2017gans,hessel2021clipscore} are widely used as auxiliary metrics to ensure that generation quality and text-image alignment remain acceptable after protection.

A third family measures \emph{residual memorization or infringement tendency}. \textbf{Memorization Score}~\cite{wen2024detecting,ren2024unveiling} is used in inference-time mitigation methods to quantify how strongly protected content is still recalled or reproduced. These metrics collectively capture the central evaluation tradeoff in Data IP behavior control: good methods should strongly suppress protected behavior without broadly harming the model's generative utility.

\paragraph{Robustness}
Robustness in this category evaluates whether constrained behavior can be recovered through adversarial prompting, inversion, or other bypass strategies. This is especially important for concept erasure, where an attacker may deliberately search for prompts or latent manipulations that reactivate the removed concept.

Two metrics are especially common. \textbf{Attack Success Rate (ASR)}~\cite{tsai2023ring,pham2023circumventing,zhang2024defensive,gong2024reliable,zhang2024generate} measures how often adversarial attacks succeed in recovering or reproducing protected content despite the defense. \textbf{Concept Recovery Rate}~\cite{tsai2023ring,pham2023circumventing,zhang2024defensive,gong2024reliable,zhang2024generate} quantifies how much of the erased concept can be recovered through such attacks. In this setting, successful suppression alone is not sufficient; a defense must also prevent reactivation of the constrained capability under adversarial attacks.

\subsubsection{Model IP Evaluation}
\label{sec:eval_behavior_model}

In the \textbf{Model IP} setting, Generative Behavior Constraint focuses on blocking capability transfer, model extraction, or unauthorized personalization. Evaluation here is still less standardized than in Data IP protection, which itself highlights a key gap in the literature.

\paragraph{Effectiveness}
A first family of metrics evaluates \emph{extraction resistance}. \textbf{Attack Model Fidelity}~\cite{hu2021model} measures the similarity between the extracted or stolen substitute model and the protected original. Lower fidelity indicates stronger protection against behavioral cloning.

A second family evaluates \emph{personalization degradation}. \textbf{Identity Score Matching}~\cite{van2023anti} measures how accurately identity is captured during unauthorized personalization, while \textbf{CLIP Similarity}~\cite{radford2021learning} is often used to assess whether protected outputs remain semantically close to the target subject or style. \textbf{ImageReward}~\cite{xu2023imagereward} provides a human-preference-aligned quality signal that is useful when protection aims to degrade the practical value of unauthorized personalized outputs rather than merely their semantic similarity.

These metrics reveal an important difference from Data IP evaluation: the goal is not only to suppress specific protected content, but to reduce the utility of an unauthorized derivative system as a commercially or functionally useful substitute.

\paragraph{Robustness}
Robustness in this category evaluates whether protection survives more adaptive extraction or personalization pipelines. \textbf{Defense Rate under DreamBooth/Textual Inversion}~\cite{ruiz2023dreambooth,gal2022image} measures whether protection remains effective when attackers use standard personalization methods. \textbf{Adaptive Attack Success Rate}~\cite{van2023anti,choi2024diffusionguard} evaluates stronger attacks specifically designed to circumvent the protection mechanism.

Robustness here is particularly underdeveloped compared with data-side unlearning. This suggests that Model IP behavior-control methods still lack a widely accepted evaluation protocol, which remains an important open problem.

\subsection{Evaluation for Attribution \& Accountability}
\label{sec:eval_attr}

Attribution \& Accountability methods are evaluated not by their ability to prevent infringement upfront, but by their ability to provide reliable post-hoc evidence. Evaluation in this category therefore emphasizes both \emph{evidentiary accuracy} and \emph{evidence persistence}. In other words, a useful attribution method must not only correctly identify ownership or misuse, but must also remain reliable after transformations, laundering attempts, or model modifications.

\subsubsection{Data IP Evaluation}
\label{sec:eval_attr_data}

In the \textbf{Data IP} setting, evaluation focuses on whether unauthorized dataset usage can be detected reliably and whether data-side watermark signals can be recovered under realistic conditions.

\paragraph{Effectiveness}
A first family of metrics evaluates \emph{detection performance}. \textbf{TPR at fixed FPR}~\cite{zhai2024membership} measures the true positive rate at a fixed false positive rate and is particularly useful when data owners need threshold-controlled forensic evidence. \textbf{AUC}~\cite{zhai2024membership} summarizes detection performance across all thresholds and is common in dataset-usage verification settings.

A second family measures \emph{direct watermark recovery}. \textbf{Detection Rate}~\cite{cui2024ftshieldwatermarkunauthorizedfinetuning,wang2024diagnosisdetectingunauthorizeddata,zhang2024attack} captures how often unauthorized usage or embedded marks are successfully identified, while \textbf{Bit Accuracy}~\cite{cui2024ftshieldwatermarkunauthorizedfinetuning,wang2024diagnosisdetectingunauthorizeddata,zhang2024attack} measures how accurately the watermark payload can be recovered. In watermarking settings, \textbf{PSNR/SSIM/LPIPS}~\cite{zhang2018unreasonable,zhang2024attack} are also used as auxiliary fidelity metrics to ensure that the embedded signal does not significantly degrade visual quality.

Together, these metrics reflect the two main evidentiary requirements of Data IP accountability: reliable detection of misuse and acceptable quality preservation of the protected content.

\paragraph{Robustness}
Robustness here asks whether detection or watermark recovery remains reliable under transformations or removal attempts. \textbf{Detection Rate/Bit Accuracy under Corruptions}~\cite{cui2024ftshieldwatermarkunauthorizedfinetuning,Zhang_2024_CVPR} measures detection performance after image corruptions such as compression, noise, and geometric modification. \textbf{Removal Attack Success Rate}~\cite{zhang2024attack} measures how often an attacker can remove or destroy the watermark while preserving image quality.

These robustness tests are crucial because evidence that disappears under mild post-processing is of limited practical value for attribution or dispute resolution.

\subsubsection{Model IP Evaluation}
\label{sec:eval_attr_model}

In the \textbf{Model IP} setting, Attribution \& Accountability evaluation focuses on whether generated content can be traced to its source model, whether model identity can be verified, and whether embedded ownership signals persist under image-level or model-level attacks.

\paragraph{Effectiveness}
A first family of metrics evaluates \emph{source attribution}. \textbf{Attribution Accuracy}~\cite{NEURIPS2023_ebb4c188,wang2024tracelatentgenerativemodel} measures the proportion of outputs correctly assigned to their generating model. This is the central metric for origin-attribution methods.

A second family evaluates \emph{fingerprint or watermark recovery}. \textbf{Bit Accuracy}~\cite{yu2022artificialfingerprintinggenerativemodels,wen2023treeringwatermarksfingerprintsdiffusion,zhao2023recipewatermarkingdiffusionmodels,fernandez2023stablesignaturerootingwatermarks,li2025diffip} measures how correctly an ownership payload is recovered from model outputs. \textbf{TPR at fixed FPR}~\cite{yu2022artificialfingerprintinggenerativemodels,wen2023treeringwatermarksfingerprintsdiffusion,zhao2023recipewatermarkingdiffusionmodels,fernandez2023stablesignaturerootingwatermarks,li2025diffip} and \textbf{p-value}~\cite{yu2022artificialfingerprintinggenerativemodels,wen2023treeringwatermarksfingerprintsdiffusion,zhao2023recipewatermarkingdiffusionmodels,fernandez2023stablesignaturerootingwatermarks,li2025diffip} are commonly used when ownership proof is framed as a detection problem with statistical confidence requirements.

A third, auxiliary metric is \textbf{FID}~\cite{heusel2017gans}, which is widely used to verify that embedding ownership evidence does not excessively degrade generation quality. This is especially important for model watermarking, where ownership proof is only practically useful if the protected model remains competitive in normal use.

\paragraph{Robustness}
Robustness in Model IP accountability focuses on evidence persistence under both \emph{image-level} and \emph{model-level} perturbations. \textbf{Bit Accuracy under Image Attacks}~\cite{jiang2023evading,fernandez2023stablesignaturerootingwatermarks,zhao2023recipewatermarkingdiffusionmodels,huang2024robin,lei2025secureefficientwatermarkinglatent} measures watermark or fingerprint recovery after transformations such as JPEG compression, cropping, resizing, blur, or adversarial image attacks. \textbf{Bit Accuracy under Model Fine-tuning/Pruning}~\cite{zhao2023recipewatermarkingdiffusionmodels,wang2025sleepermarkrobustwatermarkfinetuning,li2025diffip} evaluates whether ownership signals survive model-level modifications, including fine-tuning, pruning, or purification-style attacks.

These robustness tests are especially significant in Model IP protection because they reflect realistic laundering behavior: an attacker may not simply copy a model, but may also adapt, compress, fine-tune, or partially modify it before redeployment.

Taken together, the evaluation protocols reviewed above show that there is no single metric for IP protection in visual generative AI. Instead, meaningful evaluation depends on matching metrics to the specific risk variable a method is intended to regulate. Information Exposure Control is evaluated mainly through memorization reduction and access-security measures; Generative Behavior Constraint is evaluated through target suppression, utility retention, and resistance to behavioral bypass; and Attribution \& Accountability is evaluated through evidence accuracy and persistence. Organizing evaluation in this way makes it possible to compare methods more fairly within each protection objective, while also revealing broader gaps in the literature, such as the lack of standardized benchmarks for Model IP protection and the limited cross-architecture understanding of evaluation protocols.
\section{Regulatory Landscape and Industry Responses}
\label{sec:reg}

Technical IP protection does not operate in isolation. In practice, the deployment and effectiveness of IP safeguards for visual generative AI are increasingly shaped by laws, platform governance, licensing arrangements, provenance standards, and contractual risk allocation. From the perspective of this survey, these non-technical considerations do not map one-to-one onto a single control objective. Instead, they interact with the same underlying concerns discussed in earlier sections: limiting unauthorized data usage, constraining downstream misuse, and enabling attribution or accountability after deployment. This section therefore complements the technical taxonomy by discussing how regulation and industry practice are responding to the IP challenges of visual generative AI.

\subsection{Legal and Regulatory Approaches to IP Protection}

Across jurisdictions, legal responses to generative AI and IP are beginning to converge around four aspects: \emph{(i) human authorship and output ownership}, \emph{(ii) training-data transparency and copyright compliance}, \emph{(iii) synthetic-content disclosure and labeling}, and \emph{(iv) post-hoc accountability through enforcement, audit, or remedies}. At the same time, jurisdictions differ substantially in how they balance innovation, copyright enforcement, privacy, and platform responsibility. Table~\ref{tab:ai_law_comparison} summarizes the most relevant differences for this survey.

\paragraph{The United States}
In the United States, the regulatory and legal landscape remains largely fragmented. Copyrightability continues to be anchored in the requirement of human authorship, as reflected in \textit{Thaler v. Perlmutter} and the U.S. Copyright Office's registration guidance for works containing AI-generated material~\cite{thaler_perlmutter_dc,thaler_perlmutter_cir,usco_ai_guidance}. More recently, the U.S. Copyright Office's multi-part AI initiative has expanded this discussion through formal reports on digital replicas, copyrightability, and the use of copyrighted works in generative-AI training~\cite{usco_ai_part1_digital_replicas,usco_ai_part2_2025,usco_ai_part3_training}.

For visual generative AI, however, many of the most consequential questions remain litigation-driven rather than settled by statute. Cases such as \textit{Getty Images v. Stability AI} continue to shape debate over whether large-scale scraping and model training on copyrighted visual materials without licensing can be justified~\cite{getty_stability_complaint}. Beyond copyright, the U.S. still relies on a mix of general consumer-protection and privacy instruments, including Section 5 of the FTC Act and state-level privacy law such as the CCPA/CPRA~\cite{ftc_act_sec5,ccpa_cpra}. The U.S. model is therefore best characterized as \emph{case-driven, sectoral, and highly dynamic}, with IP norms emerging through a combination of agency interpretation, private litigation, and state-level experimentation.

\paragraph{The European Union}
The European Union has adopted the most comprehensive horizontal framework among major jurisdictions. The AI Act introduces a risk-based regulatory regime for AI systems, and its implications for generative AI extend beyond general safety concerns~\cite{eu_ai_act}. For this survey, the most relevant development is the framework governing general-purpose AI (GPAI) models, which links AI regulation directly to transparency and copyright-related obligations. In particular, GPAI providers are expected to implement policies for compliance with Union copyright law and to provide a sufficiently detailed summary of the content used for training~\cite{eu_ai_act,eu_gpai_obligations_2025,eu_gpai_code_2025,eu_gpai_template_2025}. These expectations coexist with the EU's strong baseline of personal-data protection under the GDPR~\cite{gdpr}.

Compared with the U.S., the EU approach is more \emph{ex ante}, documentation-oriented, and compliance-driven. Rather than waiting primarily for case-by-case litigation, the regulatory design places stronger weight on transparency, process obligations, and provider accountability. This is particularly relevant to the concerns of this survey, because it connects legal compliance to issues closely aligned with Information Exposure Control and Attribution \& Accountability.

\paragraph{China}
China has adopted a more top-down and service-oriented regulatory approach. Rather than centering the debate on output copyrightability, the Chinese governance has focused more heavily on provider obligations, synthetic-content management, and data governance. The \textit{Provisions on the Administration of Deep Synthesis Internet Information Services}, effective since January 2023, require providers of deep-synthesis services to strengthen management of synthetic contents and implement labeling obligations~\cite{cn_deep_synthesis}. The \textit{Interim Measures for the Administration of Generative Artificial Intelligence Services}, effective since August 2023, further extend this approach to public-facing generative-AI services, embedding expectations around lawful data use, provider responsibility, and content security~\cite{cn_genai_measures}. These AI-specific rules operate alongside the Cybersecurity Law, Data Security Law, and PIPL, which together form a broader framework for data governance, security, and personal-information protection~\cite{cn_csl,cn_dsl,cn_pipl}.

China exemplifies a regulatory strategy in which generative AI is governed largely through \emph{platform obligations and upstream service regulation}. As a result, legal control is exercised through not only the copyright doctrine, but also administrative duties imposed on AI-service providers.

\paragraph{United Kingdom}
United Kingdom occupies a distinctive position. Unlike the U.S., UK retains a statutory rule for computer-generated works under the Copyright, Designs and Patents Act 1988, which assigns authorship of computer-generated literary, dramatic, musical, or artistic works to the person making the arrangements necessary for their creation~\cite{cdpa1988}. At the same time, UK has not adopted a single AI Act analogous to the EU's. Instead, it has pursued a more flexible, regulator-led model through the AI Regulation White Paper and subsequent government responses~\cite{uk_ai_white_paper,uk_ai_white_paper_response}.

More recently, UK has become one of the most active jurisdictions in reconsidering the interface between copyright and AI training. Following the 2024 consultation on \textit{Copyright and Artificial Intelligence}, the government published a progress report in late 2025 and a fuller report and impact assessment in March 2026~\cite{uk_copyright_ai_consultation,uk_copyright_ai_progress_2025,uk_copyright_ai_report_2026}. In parallel, the Online Safety Act and subsequent policy developments around explicit deepfakes show that UK is also expanding platform and criminal-law tools for harms associated with synthetic content~\cite{uk_online_safety_act,uk_deepfake_criminalization}. UK can therefore be understood as a jurisdiction in transition: it combines an older copyright framework for computer-generated works with a rapidly evolving policy debate on AI training, transparency, and synthetic-content harms.

\begin{table*}[ht]
\centering
\caption{Comparison of legal approaches to AI-generated content and generative-AI governance across major jurisdictions.}
\label{tab:ai_law_comparison}
\resizebox*{1\linewidth}{!}{
\renewcommand{\arraystretch}{1.25}
\setlength{\tabcolsep}{3.5pt}
\begin{tabular}{p{2.3cm} p{4.35cm} p{4.35cm} p{4.35cm}}
\toprule
\textbf{Region} & \textbf{Copyright / IP Position} & \textbf{Data / Transparency Governance} & \textbf{AI-Specific Regulatory Direction} \\
\midrule
\textbf{United States} &
Human authorship remains the governing baseline for copyrightability of AI-assisted outputs; disputes over AI training on copyrighted works remain heavily litigation-driven~\cite{thaler_perlmutter_dc,thaler_perlmutter_cir,usco_ai_guidance,usco_ai_part2_2025,getty_stability_complaint}. &
No comprehensive federal AI or privacy law; governance relies on FTC enforcement, state privacy law, agency guidance, and private litigation~\cite{ftc_act_sec5,ccpa_cpra}. &
Fragmented and sectoral; strong activity around digital replicas, deepfakes, and platform accountability, but no single unified AI statute~\cite{usco_ai_part1_digital_replicas,usco_ai_part3_training}. \\
\hline

\textbf{European Union} &
Copyright remains tied to human authorship, but the AI Act directly adds copyright-related obligations for GPAI providers, including compliance policies and training-content summaries~\cite{eu_ai_act,eu_gpai_obligations_2025,eu_gpai_template_2025}. &
Strong integration with GDPR and AI Act documentation/transparency duties; public summaries of GPAI training content are part of the compliance architecture~\cite{gdpr,eu_ai_act,eu_gpai_code_2025}. &
Most comprehensive horizontal framework; combines synthetic-content transparency, GPAI obligations, systemic-risk oversight, and enforcement mechanisms~\cite{eu_ai_act,eu_gpai_code_2025}. \\
\hline

\textbf{China} &
IP law remains human-centered, while regulation of generative AI is strongly shaped by provider obligations, labeling duties, and service governance~\cite{cn_deep_synthesis,cn_genai_measures}. &
Strong emphasis on lawful data governance, personal-information protection, data security, and platform obligations under CSL, DSL, PIPL, and AI-specific service rules~\cite{cn_csl,cn_dsl,cn_pipl,cn_genai_measures}. &
Top-down governance with mandatory labeling and provider responsibility for deep-synthesis and generative-AI services~\cite{cn_deep_synthesis,cn_genai_measures}. \\
\hline

\textbf{United Kingdom} &
The UK retains statutory treatment of computer-generated works under the CDPA while actively reconsidering the copyright-training interface through consultation and formal reports~\cite{cdpa1988,uk_copyright_ai_consultation,uk_copyright_ai_progress_2025,uk_copyright_ai_report_2026}. &
Data governance follows UK GDPR and the Data Protection Act 2018; transparency and copyright issues are being revisited through AI-specific consultation rather than a single AI statute~\cite{uk_gdpr,uk_dpa2018,uk_copyright_ai_consultation}. &
Light-touch, regulator-led model; active policy development on copyright and AI, plus stronger intervention on harmful synthetic content such as explicit deepfakes~\cite{uk_ai_white_paper,uk_ai_white_paper_response,uk_online_safety_act,uk_deepfake_criminalization}. \\
\bottomrule
\end{tabular}
}
\end{table*}

Taken together, these jurisdictional differences reinforce a key insight: legal responses to generative AI rarely align perfectly with a single technical protection family. Instead, regulation typically combines asset-level concerns, transparency duties, platform obligations, and evidentiary mechanisms. This means that technical IP protection methods are increasingly likely to be evaluated not only against attack models, but also against regulatory expectations such as documentation, provenance, disclosure, and provider accountability.

\subsection{Industry Participation}

Industry participation is equally important because many operational safeguards for generative AI are being developed first as platform practices, licensing models, or interoperability standards rather than as statutory regulations. Current industry responses can be grouped into four major strategies: 1) \emph{licensed and curated data supply}, 2) \emph{provenance and disclosure}, 3) \emph{creator controls and platform enforcement}, and 4) \emph{contractual risk allocation}. These strategies do not map perfectly onto the Control Logic View, but they interact closely with the same practical concerns of exposure, behavior, and accountability.

\paragraph{Licensed and Curated Data Supply}
One major industry practice is to reduce upstream legal uncertainty by shifting from broad web scraping toward licensed, curated, or provenance-controlled data pipelines. Adobe has explicitly positioned Firefly around this logic, stating that Firefly models are trained on licensed content, such as Adobe Stock, together with openly licensed or public-domain content, and that customers' private content is not used for Firefly training by default~\cite{adobe_firefly_approach}. Shutterstock has adopted a complementary licensing-centered strategy by expanding its partnership with OpenAI through a six-year agreement to provide image, video, music, and metadata resources as high-quality training data for generative-AI models~\cite{shutterstock_openai_licensing}. Getty Images has pursued a more product-facing version of the same logic by launching a commercially positioned generative-AI offering trained on its licensed creative content, with usage rights, legal protection, and contributor compensation framed as part of its product differentiation~\cite{getty_generative_ai_launch}. Together, these cases show how content licensing, dataset provenance, and compensation mechanisms are becoming practical instruments for mitigating Data IP risks in visual generative AI.

These arrangements matter because they complement technical information-exposure controls with a market mechanism: instead of only trying to prevent unauthorized learning after the fact, firms increasingly seek to secure lawful data access in advance.

\paragraph{Provenance, Disclosure, and Traceability}
Another industry practice focuses on post-hoc transparency and content provenance. OpenAI states that images generated in ChatGPT can include C2PA metadata, an open standard for attaching provenance information to media~\cite{openai_c2pa_help,c2pa_spec}. Adobe has advanced a broader provenance ecosystem through Content Credentials, which attach standardized metadata about how content was created or edited, including whether AI tools were used~\cite{adobe_content_credentials}. Google DeepMind has pursued a closely related strategy through SynthID, which is designed to watermark and identify AI-generated content~\cite{google_synthid,google_synthid_detector}.

These efforts are especially relevant to the \textit{Attribution \& Accountability} branch of this survey. They also illustrate an important trend: provenance tooling is becoming an interoperability layer across platforms, model providers, and downstream distribution channels, rather than a purely model-internal defense.

\paragraph{Creator Controls and Platform Enforcement}
The third industry practice centers on creator-facing controls, notice-and-takedown processes, and platform governance. Midjourney's terms include a DMCA workflow for removal and counter-notification, embedding copyright dispute handling into platform operations~\cite{midjourney_terms}. OpenAI has publicly described \textit{Media Manager} as a planned tool intended to let creators and content owners specify how their works should be included or excluded from machine-learning research and training~\cite{openai_data_ai_media_manager}. These mechanisms are still uneven across providers, and their practical coverage is far from standardized. Nevertheless, they indicate that platform-level creator control is becoming an increasingly important complement to both legal enforcement and technical protection.

\paragraph{Contractual Risk Allocation and Enterprise Assurance}
The fourth industry practice involves contractual allocation of IP risk, especially for enterprise customers. OpenAI's Services Agreement and Service Terms include IP indemnity commitments for enterprise use of outputs in certain circumstances~\cite{openai_services_agreement,openai_service_terms_output_indemnity}. Adobe likewise markets contractual IP indemnification for eligible Firefly enterprise offers~\cite{adobe_firefly_indemnity_faq}. These commitments do not eliminate upstream copyright disputes, but they matter commercially as they convert part of the residual IP risk into a contractual assurance that can support adoption in professional workflows.

Overall, regulatory and industry developments reveal that the governance of visual generative AI is moving towards a layered model. Public regulation increasingly emphasizes transparency, disclosure, lawful data governance and accountability, while industry practice increasingly emphasizes licensed data, provenance standards, creator-control channels and contractual risk management. These developments reinforce a central point: technical IP protection methods are most useful when they can interoperate with external governance frameworks, rather than being treated as isolated model-level safeguards.
\section{Open Challenges and Future Directions}
\label{sec:future}

Despite the rapid progress reviewed in this survey, IP protection for visual generative models is still at an early stage. Existing methods remain fragmented across different threat assumptions, model architectures, and deployment scenarios, and several core research problems are still insufficiently addressed. Below, we highlight the most important open challenges and future directions.

\paragraph{Standardized Benchmarks and Evaluation Protocols}
A major limitation of the current literature is the lack of standardized benchmarks and evaluation protocols. As discussed in Section~\ref{sec:eval}, different works often rely on different datasets, target concepts, attack settings, and utility metrics, making cross-paper comparison difficult. This problem is especially severe for Model IP protection, where evaluation is often borrowed from adjacent tasks rather than designed specifically for IP risks. Future work should therefore establish benchmark suites aligned with the three control objectives in this survey, including standardized threat models, shared data splits, unified utility metrics, and adaptive attack protocols. In particular, benchmark construction should go beyond static effectiveness measurement and explicitly include robustness under prompt attacks, concept recovery, watermark removal, model purification, and fine-tuning-based bypass.

\paragraph{Proactive Protection for Model IP}
Compared with Data IP protection, proactive protection for Model IP remains significantly underexplored. Most existing methods for Model IP are reactive, focusing on watermarking, fingerprinting, or origin attribution after a model has already been reused or redistributed. In contrast, relatively few works directly prevent unauthorized access to proprietary model value before infringement occurs. This gap is likely to become more serious as model APIs, downstream fine-tuning, and model customization become increasingly common. Future work should therefore place greater emphasis on proactive Model IP defenses, including stronger model locking, anti-extraction mechanisms for black-box APIs, capability-gated generation, and ownership-preserving fine-tuning schemes. More broadly, the field needs clearer threat models that distinguish between checkpoint theft, behavioral cloning, derivative adaptation, and capability laundering.

\paragraph{Precise and Explainable Evidence for Reactive Protection}
Reactive protection methods are only practically useful if they can provide evidence that is not only accurate, but also interpretable and actionable. Current attribution and detection methods often produce coarse binary judgments, such as whether a dataset was likely used or whether an image was likely generated by a certain model, but they rarely explain which part of the output, which training signal, or which model component supports this conclusion. This limits their value in legal disputes, platform moderation, and creator-facing enforcement. Future work should therefore move toward localized and explainable IP evidence, such as identifying copied regions, style-bearing features, concept-specific activation patterns, or source-dependent provenance traces. Such methods would make reactive protection substantially more useful for copyright claims, takedown requests, and audit procedures.

\paragraph{Robustness Under Adaptive Attacks and Post-Protection Adaptation}
Robustness remains one of the weakest aspects of the current literature. Many methods are still evaluated under relatively simple attack settings, even though realistic adversaries can iteratively refine prompts, use inversion to recover erased concepts, fine-tune away watermarks, or distill protected behavior into a substitute model. In addition, many protection methods are not tested under post-protection adaptation, such as continued fine-tuning, personalization, or domain transfer. Future work should therefore treat adaptive robustness as a first-class objective rather than a secondary stress test. This includes stronger red-teaming for concept recovery, systematic evaluation of model laundering and watermark removal, and explicit study of whether protected or erased capabilities re-emerge after subsequent adaptation.

\paragraph{Beyond Diffusion-Centric Protection}
Most current IP protection studies are developed and evaluated primarily on diffusion models. While this focus is understandable given the dominance of diffusion in current visual generation, it leaves open a fundamental question: which protection mechanisms are genuinely general, and which are specific to diffusion architectures? Autoregressive models, flow-based models, multimodal native generators, and interactive world models expose different internal interfaces and may exhibit different forms of memorization, extraction, attribution, or style imitation. Future work should therefore investigate IP protection beyond diffusion and clarify which defenses transfer across architectures. This is important not only for broader coverage, but also for identifying architecture-aware protection mechanisms that are better matched to the actual structure of the underlying generator.

\paragraph{From Single-Modality Protection to Multimodal and Personalized Systems}
Another important limitation of the current literature is its strong focus on image generation in relatively static settings. In practice, generative systems are rapidly becoming multimodal, personalized, and continuously updated. Future models increasingly combine image, video, audio, text, and editing capabilities, while personalization pipelines enable users to adapt a base model to specific identities, styles, or domains. These developments create new forms of IP risk, because protected content can be transferred across modalities, amplified through personalization, or reintroduced during continual adaptation. Future work should therefore study whether current protection methods remain effective in multimodal and personalized settings, and whether new forms of protection are required for cross-modal transfer, subject-driven generation, and continual reuse.

\paragraph{Integration with Provenance Standards, Platform Governance, and Regulation}
Finally, future technical work should be designed with deployment and governance constraints in mind. As discussed in Section~\ref{sec:reg}, regulation and industry practice increasingly emphasize transparency, provenance, creator control, and accountability. However, many technical methods are still developed in isolation from these external requirements. Future work should therefore better connect technical IP protection with content credentials, provenance metadata, platform-level enforcement workflows, creator opt-out or licensing mechanisms, and emerging regulatory obligations on training-data disclosure and synthetic-content labeling. In other words, an effective future protection system will likely need to function not as a standalone algorithm, but as part of a broader governance framework spanning model developers, platforms, creators, auditors, and regulators.

Overall, a central challenge for the next stage of this field is to move from isolated protection techniques toward a more systematic framework that jointly considers threat modeling, benchmark design, proactive defense, reactive evidence, robustness, architectural diversity, and deployment compatibility. Advancing along these directions will be critical for building IP-aware generative models that are not only technically effective, but also practically enforceable.
\section{Conclusion}
\label{sec:concl}

In this survey, we presented a structured review of IP protection for visual generative AI through a \textit{Control Logic View}. Rather than organizing the literature by lifecycle stage or technical mechanism alone, we argued that IP protection methods are more systematically understood by the specific risk variable they regulate. Under this perspective, we unified existing work into three control objectives: \textit{Information Exposure Control}, \textit{Generative Behavior Constraint}, and \textit{Attribution \& Accountability}. Within each objective, we further distinguished between \textit{Data IP} and \textit{Model IP} protection, thereby providing a common framework for understanding what is being protected, how infringement risk arises, and where technical interventions are applied.

Based on this taxonomy, we reviewed the main families of protection methods developed for visual generative models, ranging from data filtering, perturbation, and compositional training to concept unlearning, safety-aligned inference, watermarking, fingerprinting, and origin attribution. We also showed that evaluation should not be treated as an independent or purely metric-driven component, but should instead be organized under the same control logic as the protection methods themselves. This view helps clarify why different categories of methods require different notions of effectiveness and robustness, and why benchmark design remains one of the central open problems in the field.

Beyond technical methods, we further highlighted that IP protection for visual generative AI is increasingly shaped by regulatory developments and industry practice. Emerging governance frameworks across major jurisdictions are placing growing emphasis on training-data transparency, synthetic-content disclosure, provenance, and provider accountability, while industry actors are responding through licensed data pipelines, creator-facing controls, watermarking, provenance standards, and contractual assurance. These developments suggest that future IP protection will likely depend not on isolated model-level defenses alone, but on the interaction between technical mechanisms, platform governance, legal compliance, and deployment practice.

Overall, the current landscape reveals both substantial progress and significant fragmentation. Existing methods have demonstrated that exposure reduction, behavior control, and post-hoc attribution are all viable components of an IP protection toolbox, yet the field still lacks standardized benchmarks, strong proactive Model IP defenses, broadly applicable cross-architecture methods, and evidence mechanisms that are sufficiently robust, interpretable, and operational in real-world settings. We hope that the taxonomy and synthesis provided in this survey can serve as a useful foundation for future research toward more principled, effective, and deployable IP protection for visual generative AI.

\bibliographystyle{ACM-Reference-Format}
\bibliography{mainbib}

\end{document}